\documentclass[runningheads]{llncs}
\usepackage{graphicx}

\usepackage{tikz}
\usepackage{comment}
\usepackage{amsmath,amssymb} 
\usepackage{color}

\usepackage[accsupp]{axessibility}  

\usepackage{multirow}
\usepackage{bbold}
\usepackage{caption}
\usepackage{mathtools}
\usepackage{amsfonts}
\usepackage{booktabs}
\usepackage{etoolbox}
\usepackage{graphicx}
\usepackage{soul}
\usepackage{color}
\usepackage{xspace}
\usepackage{subcaption}
\usepackage{amsmath}
\usepackage{algorithm}
\usepackage{algpseudocode}
\usepackage{mathtools}
\usepackage{siunitx}
\usepackage{xcolor}
\usepackage{hyperref}
\usepackage{CJKutf8}

\usepackage{amssymb}
\usepackage{pifont}
\usepackage{wrapfig}
\newcommand{\cmark}{\ding{51}}%
\newcommand{\xmark}{\ding{55}}%
\usepackage[normalem]{ulem}
\newcommand{\quadd}{\hspace{0.5em}}
\newcommand{\quaddd}{\hspace{0.25em}}

\def\etal{\emph{et~al.}}

\definecolor{Highlight}{HTML}{39b54a}  
\definecolor{green}{HTML}{39b54a}
\definecolor{red}{HTML}{cb4335}
\definecolor{rowcolor}{RGB}{209,154,128}

\AtBeginEnvironment{tabular}{\scriptsize}

\begin{document}
\pagestyle{headings}
\mainmatter
\title{Making Heads or Tails: Towards Semantically Consistent Visual Counterfactuals}

\titlerunning{Making heads or tails}
\author{
    Simon Vandenhende \quad
    Dhruv Mahajan \\
    Filip Radenovic\thanks{Equal contribution.} \quad
    Deepti Ghadiyaram$^{\star}$
}
\institute{Meta AI}
\authorrunning{Vandenhende et al.}


\maketitle
\begin{abstract}
A visual counterfactual explanation replaces image regions in a query image with regions from a distractor image such that the system's decision on the transformed image changes to the distractor class. In this work, we present a novel framework for computing visual counterfactual explanations based on two key ideas. First, we enforce that the \textit{replaced} and \textit{replacer} regions contain the same semantic part, resulting in more semantically consistent explanations. Second, we use multiple distractor images in a computationally efficient way and obtain more discriminative explanations with fewer region replacements. Our approach is $\mathbf{27\%}$ more semantically consistent and an order of magnitude faster than a competing method on three fine-grained image recognition datasets. We highlight the utility of our counterfactuals over existing works through machine teaching experiments where we teach humans to classify different bird species. We also complement our explanations with the vocabulary of parts and attributes that contributed the most to the system's decision. In this task as well, we obtain state-of-the-art results when using our counterfactual explanations relative to existing works, reinforcing the importance of semantically consistent explanations. Source code is available at \href{https://github.com/facebookresearch/visual-counterfactuals}{github.com/facebookresearch/visual-counterfactuals}.
\end{abstract}

\section{Introduction}
\label{sec: introduction}
Explainable AI (XAI) research aims to develop tools that allow lay-users to comprehend the reasoning behind an AI system's decisions~\cite{markus2021role,zablocki2021explainability}. XAI tools are critical given the pervasiveness of computer vision technologies in various human-centric applications such as self-driving vehicles, healthcare systems, and facial recognition tools. These tools serve several purposes~\cite{adadi2018peeking,wachter2018counterfactual}: (i) they help users understand why a decision was reached thereby making systems more transparent, (ii) they allow system developers to improve their system, and (iii) they offer agency to users affected by the system's decision to change the outcome.

\begin{figure}[t]
     \centering
     \includegraphics[width=0.98\textwidth]{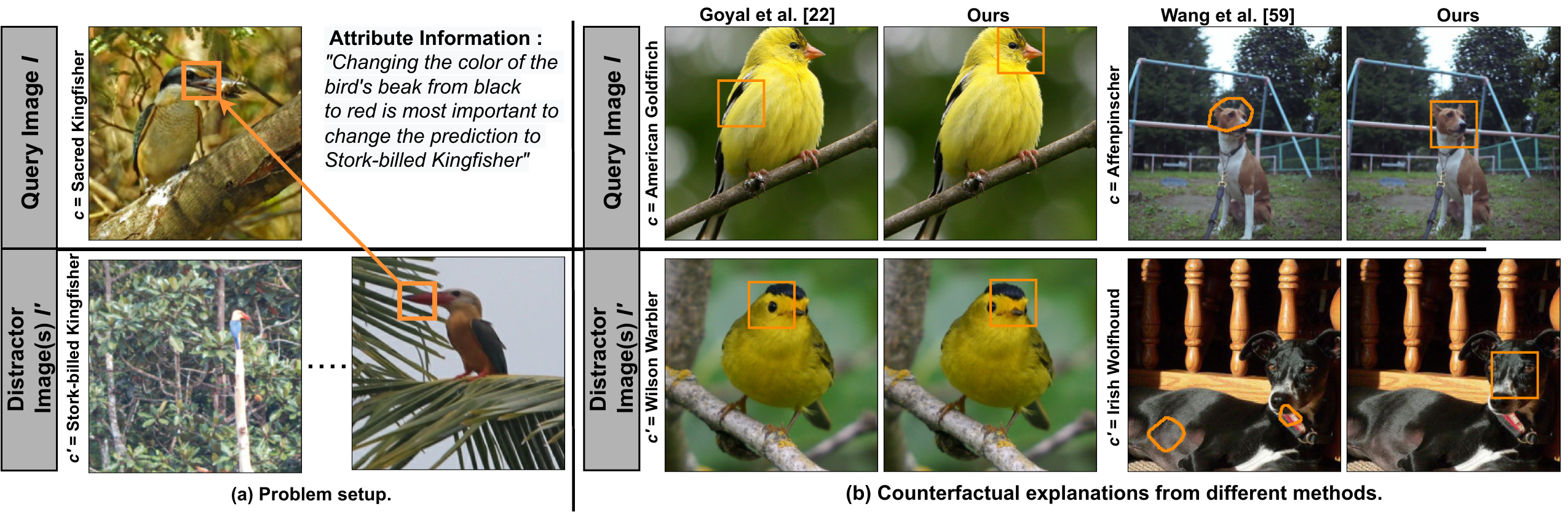}
     \vspace*{-1.0em}
     \caption{\footnotesize{\textbf{Paper overview. (a)} Given a query image $I$ (top row) from class $c$, we provide counterfactual explanations relative to a distractor image $I'$ (bottom row) from class $c'$. The explanations highlight what regions in $I$ should be replaced from $I'$ for the transformed image to be classified as $c'$. We also use attribute information to identify the region attributes that contributed the most for a counterfactual. \textbf{(b)} Unlike~\cite{goyal2019counterfactual,wang2020scout}, our explanations identify regions that are both discriminative and semantically similar.}}
     \label{fig: teaser}
     \vspace{-1.2em}
\end{figure}
One intuitive way to explain a system's decision is through counterfactual explanations~\cite{verma2020counterfactual,wachter2018counterfactual} which describe \textit{in what way} a data instance would need to be different in order for the system to reach an \textit{alternate} conclusion. In this work, we study counterfactual explanations for fine-grained image recognition tasks, where the most confusing classes are often hard to distinguish. The difficulty of this problem makes it a particularly well suited setting to study intuitive and human-understandable explanations. 
Figure~\ref{fig: teaser}-a presents a \emph{query image} $I$ and a \emph{distractor image} $I'$ belonging to the categories \textit{Sacred Kingfisher} ($c$) and \textit{Stork-billed Kingfisher} $(c')$, respectively. Given a black-box classification model, a counterfactual explanation aims to answer: ``how should the query image $I$ change for the model to predict $c'$ instead of $c$?'' To do this, we utilize the distractor image $I'$ (or a set of distractor images) and identify which regions in $I$ should be replaced with regions from $I'$ for the model's prediction to be $c'$.

Counterfactual visual explanations are under-explored~\cite{goyal2019counterfactual,wang2020scout}, and most popular XAI methods use saliency maps~\cite{dabkowski2017real,fong2017interpretable,petsiuk2018rise,selvaraju2017grad,zhou2016learning} or feature importance scores~\cite{datta2016algorithmic,kim2018interpretability,lundberg2017unified,ribeiro2016should,ribeiro2018anchors,sundararajan2017axiomatic,zintgraf2017visualizing} to highlight what image regions or features most contribute to a model's decision. Unlike counterfactual explanations, these methods do not consider alternate scenarios which yield a different result. Additionally, some of these methods~\cite{lundberg2017unified,ribeiro2016should,ribeiro2018anchors} extract explanations via a local model approximation, leading to explanations that are \textit{unfaithful}~\cite{adebayo2018sanity,slack2020fooling}, i.e., they misrepresent the model's behavior. By contrast, current counterfactual explanations are faithful by design as they operate on the original model's output to generate explanations. Further, counterfactuals share similarities with how children learn about a concept -- by contrasting with other related concepts~\cite{beck2009relating,buchsbaum2012power}. As studied in~\cite{miller2019explanation,verma2020counterfactual,wachter2018counterfactual}, an ideal counterfactual should have the following properties: (i) the highlighted regions in the images $I, I'$ should be \underline{discriminative} of their respective classes; (ii) the counterfactual should be sensible in that the replaced regions should be \underline{semantically consistent}, i.e., they correspond to the same object parts; and, (iii) the counterfactual should make as few changes as possible to the query image $I$ as humans find sparse explanations \underline{easier to understand}.

Prior works~\cite{goyal2019counterfactual,wang2020scout} proposed ways to identify the most discriminative image regions to generate counterfactual explanations. However, naively applying this principle can yield degenerate solutions that are semantically inconsistent. Figure~\ref{fig: teaser}-b visualizes such scenarios, where prior works~\cite{goyal2019counterfactual,wang2020scout} replace image regions corresponding to different object parts (e.g.,~\cite{goyal2019counterfactual} replaces bird's wing in $I$ with a head in $I'$). Further, these methods rely on a single distractor image $I'$, which often limits the variety of discriminative regions to choose from, leading to explanations that are sometimes less discriminative hence uninformative.

This paper addresses these shortcomings. Specifically, we propose a novel and computationally efficient framework that produces both discriminative and semantically consistent counterfactuals. Our method builds on two key ideas. First, we constrain the identified class-specific image regions that alter a model's decision to allude to the same semantic parts, yielding more semantically consistent explanations. Since we only have access to object category labels, we impose this as a soft constraint in a separate auxiliary feature space learned in a self-supervised way. Second, contrary to prior works, we expand the search space by using multiple distractor images from a given class leading to more discriminative explanations with fewer regions to replace. However, naively extending to multiple distractor images poses a computational bottleneck. We address this by constraining the processing to only the most similar regions by once again leveraging the soft constraint, resulting in an order of magnitude speedup.

Our approach significantly outperforms the s-o-t-a~\cite{goyal2019counterfactual,wang2020scout} across several metrics on three datasets -- CUB~\cite{CUB}, Stanford-Dogs~\cite{DOGS}, and iNaturalist-2021~\cite{INATURALIST} and yields more semantically consistent counterfactuals (Fig.~\ref{fig: teaser}-b). While prior work~\cite{goyal2019counterfactual} suffers computationally when increasing the number of distractor images, the optimization improvements introduced in our method make it notably efficient. We also study the properties of the auxiliary feature space and justify our design choices. Further, we show the importance of generating semantically consistent counterfactuals via a machine teaching task where we teach lay-humans to recognize bird species. We find that humans perform better when provided with our semantically consistent explanations relative to others~\cite{goyal2019counterfactual,wang2020scout}. 

We further reinforce the importance of semantically consistent counterfactuals by proposing a method to complement our explanations with the vocabulary of parts and attributes. Consider Fig.~\ref{fig: teaser}-a, where the counterfactual changes both the color of the beak and forehead. Under this setup, we provide nameable parts and attributes corresponding to the selected image regions and inform what attributes contributed the most to the model's decision. For example, in Fig.~\ref{fig: teaser}-a, our explanation highlights that the beak's color mattered the most. We find that our explanations identify class discriminative attributes -- those that belong to class $c$ but not to $c'$, or vice versa -- and are more interpretable. 

In summary, our contributions are: \textbf{(i)} we present a framework to compute semantically consistent and faithful counterfactual explanations by enforcing the model to only replace semantically matching image regions (Sec.~\ref{subsec: method_consistency}), \textbf{(ii)} we leverage multiple distractor images in a computationally efficient way, achieve an order of magnitude speedup, and generate more discriminative and sparse explanations (Sec.~\ref{subsec: method_multi_distractor}), \textbf{(iii)} we highlight the utility of our framework through extensive experiments (Sec.~\ref{subsec: sota} -~\ref{subsec: ablation}) and a human-in-the-loop evaluation through machine teaching (Sec.~\ref{subsec: machine_teaching}), \textbf{(iv)} we augment visual counterfactuals with nameable part and attribute information (Sec.~\ref{sec: attributes}).
\section{Related Work}
\label{sec: related_work}
\noindent\textbf{Feature attribution methods}~\cite{ancona2018towards} rely on the back propagation algorithm \cite{bach2015pixel,rebuffi2020there,selvaraju2017grad,shrikumar2016not,simonyan2013deep,zeiler2014visualizing,zhou2016learning} or input perturbations~\cite{chang2018explaining,dabkowski2017real,dhurandhar2018explanations,fong2017interpretable,petsiuk2018rise,zintgraf2017visualizing} to identify the image regions that are most important to a model's decision. However, none of these methods can tell how the image should change to get a different outcome. 

\noindent\textbf{Counterfactual explanations}~\cite{mothilal2020explaining,poyiadzi2020face,verma2020counterfactual,wachter2018counterfactual} transform a query image $I$ of class $c$ such that the model predicts class $c'$ on the transformed image. In computer vision, several works~\cite{alipour2021improving,hvilshoj2021ecinn,jacob2021steex,lang2021explaining,liu2019generative,rodriguez2021beyond,singla2019explanation,singla2021explaining} used a generative model to synthesize counterfactual examples. However, the difficulties of realistic image synthesis can limit these methods~\cite{hvilshoj2021ecinn,liu2019generative,rodriguez2021beyond,singla2019explanation} to small-scale problems. A few works~\cite{alipour2021improving,jacob2021steex,singla2021explaining} guided the image generation process via pixel-level supervision to tackle more complex scenes. StyleEx~\cite{lang2021explaining} uses the latent space of a StyleGAN~\cite{karras2020analyzing} to identify the visual attributes that underlie the classifier's decision. Despite these efforts, it remains challenging to synthesize realistic counterfactual examples. Our method does not use a generative model but is more related to the works discussed next.

A second group of works~\cite{akula2020cocox,goyal2019counterfactual,wang2020scout} finds the regions or concepts in $I$ that should be changed to get a different outcome. CoCoX~\cite{akula2020cocox} identifies visual concepts to add or remove to change the prediction. Still, the most popular methods~\cite{goyal2019counterfactual,wang2020scout} use a distractor image $I'$ from class $c'$ to find and replace the regions in $I$ that change the model's prediction to $c'$. SCOUT~\cite{wang2020scout} finds these regions via attribute maps. Goyal~\etal~\cite{goyal2019counterfactual} use spatial features of the images to construct counterfactuals. These methods have two key advantages. First, the distractor images are often readily available and thus inexpensive to obtain compared to pixel-level annotations~\cite{alipour2021improving,jacob2021steex,singla2021explaining}. Second, these methods fit well with fine-grained recognition tasks, as they can easily identify the distinguishing elements between classes. Our framework follows a similar strategy but differs in two crucial components. First, we enforce that the replaced regions are semantically consistent. Second, our method leverages multiple distractor images in an efficient way.
\section{Method}
\label{sec: method}
\begin{figure}[t]
    \centering
    \includegraphics[width=0.85\textwidth]{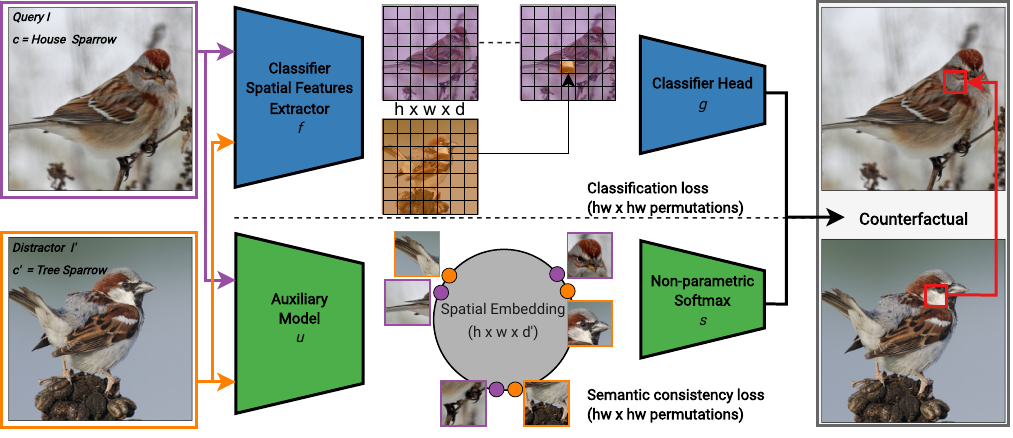}
    \vspace{-1.0em}
    \caption{\footnotesize{\textbf{Our counterfactual explanation} identifies regions in a query image $I$ from class $c$ and a distractor image $I'$ from class $c'$ such that replacing the regions in $I$ with the regions in $I'$ changes the model's outcome to $c'$. Instead of considering actual image regions, we operate on $h \times w$ cells in the spatial feature maps. The cells are selected based upon: (i) a classification loss that increases the predicted probability $g_{c'}$ of class $c'$ and (ii) a semantic consistency loss that selects cells containing the same semantic parts. We use a self-supervised auxiliary model to compute the semantic loss.}}
    \label{fig: method}
\end{figure}

Our key goal is to: (i) generate a counterfactual that selects discriminative and semantically consistent regions in $I$ and $I'$ without using additional annotations, (ii) leverage multiple distractor images efficiently. We first review the foundational method~\cite{goyal2019counterfactual} for counterfactual generation that our framework builds on and then introduce our approach, illustrated in Fig.~\ref{fig: method}.

\subsection{Counterfactual problem formulation: preliminaries} 
\label{subsec: method_preliminaries} 
Consider a deep neural network with two components: a spatial feature extractor $f$ and a decision network $g$. Note that any neural network can be divided into such components by selecting an arbitrary layer to split at. In our setup, we split a network after the final down-sampling layer. The spatial feature extractor $f:\mathcal{I} \to \mathbb{R}^{hw \times d}$ maps the image to a $h \times w \times d$ dimensional spatial feature, reshaped to a $hw \times d$ spatial cell matrix, where $h$ and $w$ denote the spatial dimensions and $d$ the number of channels. The decision network $g: \mathbb{R}^{hw \times d} \to \mathbb{R}^{|\mathcal{C}|}$ takes the spatial cells and predicts probabilities over the output space $\mathcal{C}$. Further, let query and distractor image $I, I' \in \mathcal{I}$ with class predictions $c, c' \in \mathcal{C}$.

Following~\cite{goyal2019counterfactual}, we construct a counterfactual $I^*$ in the feature space $f(.)$ by replacing spatial cells in $f(I)$ with cells from $f(I')$ such that the classifier predicts $c'$ for $I^*$. This is done by first rearranging the cells in $f(I')$ to align with $f(I)$ using a permutation matrix $P \in \mathbb{R}^{hw \times hw}$, then selectively replacing entries in $f(I)$ according to a sparse binary gating vector $\mathbf{a} \in \mathbb{R}^{hw}$. Let $\circ$ denote the Hadamard product. The transformed image $I^*$ can be written as:
\vspace{-0.4em}
\begin{equation}
    f(I^*) = (\mathbb{1} - \mathbf{a}) \circ f(I) + \mathbf{a} \circ P f(I')
\vspace{-0.4em}
\end{equation}
\noindent \textbf{Classification loss:} Recall that our first goal is to identify class-specific image regions in $I$ and $I'$ such that replacing the regions in $I$ with those in $I'$ increases the predicted probability $g_{c'}(.)$ of class $c'$ for $I^*$. To avoid a trivial solution where all cells of $I$ are replaced, a sparsity constraint is applied on $\mathbf{a}$ to minimize the number of cell edits ($m$). Following the greedy approach from~\cite{goyal2019counterfactual}, we iteratively replace spatial cells in $I$ by repeatedly solving Eq.~\ref{eq: class_loss} that maximizes the predicted probability $g_{c'}(\cdot)$ until the model's decision changes. 
\vspace{-0.4em}
\begin{equation}
    \label{eq: class_loss}
    \max_{P, a} g_{c'}((\mathbb{1} - \mathbf{a}) \circ f(I) + \mathbf{a} \circ P f(I')) \text{ with } ||\mathbf{a}||_1 = 1 \text{ and } a_i \in \left\{0,1\right\}
\vspace{-0.4em}
\end{equation}
We evaluate $g_{c'}$ for each of the $h^2w^2$ permutations constructed by replacing a single cell in $f(I)$ with an arbitrary cell in $f(I')$. The computational complexity is $2 \cdot C_f + m h^2 w^2 \cdot C_g$, where $C_f$ and $C_g$ denote the cost of $f$ and $g$ respectively.

Eq.~\ref{eq: class_loss} does not guarantee that the replaced cells are semantically similar. For example, in the task of bird classification, the counterfactual could replace the wing in $I$ with head in $I'$ (e.g., Fig.~\ref{fig: teaser}-b) leading to nonsensical explanations. We address this problem via a semantic consistency constraint, described next.

\subsection{Counterfactuals with a semantic consistency constraint}
\label{subsec: method_consistency}
Consider an embedding model $u: \mathcal{I} \to \mathbb{R}^{hw \times d'}$ that brings together spatial cells belonging to the same semantic parts and separates dissimilar cells. Let $u(I)_i$ denote the feature of the $i$-th cell in $I$. We estimate the likelihood that cell $i$ of $I$ semantically matches with cell $j$ of $I'$ by:
\vspace{-0.4em}
\begin{equation}
\label{eq: similarity_loss}
\mathcal{L}_s(u(I)_i, u(I')_{j}) = \frac{\exp (u(I)_i \cdot u(I')_{j} / \tau)}{\sum_{j' \in u(I')} \exp(u(I)_i \cdot u(I')_{j'} / \tau)},
\vspace{-0.4em}
\end{equation}
where $\tau$ is a temperature hyper-parameter that relaxes the dot product. Eq.~\ref{eq: similarity_loss} estimates a probability distribution of a given query cell $i$ over all distractor cells $j'$ using a non-parametric softmax function and indicates what distractor cells are most likely to contain semantically similar regions as the query cell $i$. Like the classification loss (Eq.~\ref{eq: class_loss}), we compute the semantic loss for all $h^2w^2$ cell permutations. Thus, the complexity is $2 \cdot C_u + h^2w^2 \cdot C_{\text{dot}}$, where $C_u, C_{\text{dot}}$ denote the cost of the auxiliary model $u$ and the dot-product operation respectively. Empirically, we observe that dot-products are very fast to compute and the semantic loss adds a tiny overhead to the overall computation time. Note that unlike the classification loss which is computed for each edit, $\mathcal{L}_s$ is computed only once in practice, i.e., the cost gets amortized for multiple edits.\vspace*{0.05in}\\
\noindent\textbf{Total loss:} 
We combine both losses to find the single best cell edit:
\vspace{-0.4em}
\begin{equation}
    \label{eq: total_loss}
    \begin{split}
    \max_{P, a} \quadd \log \quaddd \underbrace{g_{c'}((\mathbb{1} - \mathbf{a}) \circ f(I) + \mathbf{a} \circ P f(I'))}_{\text{Classification loss } \mathcal{L}_c} \quaddd + \quaddd \lambda \cdot \log \underbrace{\mathcal{L}_s(\mathbf{a}^Tu(I), \mathbf{a}^T P u(I'))}_{\text{Semantic consistency loss } \mathcal{L}_s} \\ 
    \text{with $P \in \mathbb{R}^{hw\times hw}$, } ||\mathbf{a}||_1 = 1 \text{ and } a_i \in \left\{0,1\right\}, \text{ and } \lambda \text{ balances } \mathcal{L}_c \text{ and }\mathcal{L}_s.
\end{split}
\vspace{-0.4em}
\end{equation} 
We reiterate that $\mathcal{L}_c$ optimizes to find class-specific regions while $\mathcal{L}_s$ ensures that these regions semantically match. We also stress that our explanations are faithful with respect to the underlying deep neural network, since, the proposed auxiliary model, irrespective of the value of $\lambda$, only acts as a regularizer and does not affect the class predictions of the transformed images.\vspace*{0.05in}\\
\noindent\textbf{Choice of auxiliary model: } An obvious choice is to use the spatial feature extractor $f$ as the auxiliary model $u$. We empirically found that since $f$ is optimized for an object classification task, it results in an embedding space that often separates instances of similar semantic parts and is thus unfit to model region similarity. We found that self-supervised models are more appropriate as auxiliary models for two reasons: a) they eliminate the need for part location information, b) several recent studies~\cite{caron2021emerging,van2021revisiting,van2021unsupervised} showed that self-supervised models based on contrastive learning~\cite{chen2020simple,he2020momentum,wu2018unsupervised} or clustering~\cite{asano2019self,caron2018deep,caron2020unsupervised,van2020scan} learn richer representations that capture the semantic similarity between local image regions as opposed to task-related similarity in a supervised setup. Such representations have been valuable for tasks such as semantic segment retrieval~\cite{van2021revisiting}. Thus, the resulting embedding space inherently brings together spatial cells belonging to the same semantic parts and separates dissimilar cells (see Table~\ref{tab: auxiliary_model}).

\subsection{Using multiple distractor images through a semantic constraint}
\label{subsec: method_multi_distractor}
Recall, the method uses spatial cells from $f(I')$ to iteratively construct $f(I^*)$. Thus, the quality of the counterfactual is sensitive to the chosen distractor image $I'$. Having to select regions from a single distractor image can limit the variety of discriminative parts to choose from due to factors like pose, scale, and occlusions. We address this limitation by leveraging multiple distractor images from class $c'$. In this way, we expand our search space in Eq.~\ref{eq: total_loss}, allowing us to find highly discriminative regions that semantically match, while requiring fewer edits.

However, leveraging ($n$) multiple distractor images efficiently is not straight-forward as it poses a significant computational overhead. This is because, in this new setup, for each edit we can pick any of $n \times hw$ cells from the $n$ distractor images. This makes the spatial cell matrix of the distractor images of shape $nhw\times d$, the matrix $P$ $hw \times nhw$, and $\mathbf{a} \in \mathbb{R}^{hw}$. $\mathcal{L}_c$ (Eq.~\ref{eq: class_loss}) with a single distractor image is already expensive to evaluate due to: (i) its quadratic dependence on $hw$ making the cell edits memory intensive and, (ii) the relatively high cost of evaluating $g$, involving at least one fully-connected plus zero or more conv layers. This computation gets amplified by a factor $n$ with multiple distractor images.

On the other hand, $\mathcal{L}_s$ (Eq.~\ref{eq: similarity_loss}) is computationally efficient as: (i) it does not involve replacing cells and (ii) the dot-product is inexpensive to evaluate. Thus, we first compute $\mathcal{L}_s$ (Eq.~\ref{eq: similarity_loss}) to select the top-$k\%$ cell permutations with the lowest loss, excluding the ones that are likely to replace semantically dissimilar cells. Next, we compute $\mathcal{L}_c$ (Eq.~\ref{eq: class_loss}) only on these selected top-$k\%$ permutations. With this simple trick, we get a significant overall speedup by a factor $k$ (detailed analysis in suppl.). Thus, our overall framework leverages richer information, produces semantically consistent counterfactuals, and is about an order of magnitude faster than~\cite{goyal2019counterfactual}. Note that the multi-distractor setup can be extended to~\cite{goyal2019counterfactual} but not to SCOUT~\cite{wang2020scout}, as the latter was designed for image pairs.
\section{Experiments}
\label{sec: experiments}

\newcommand{\reshg}[2]{{#1} \tiny{\textcolor{green}{($+{#2}$)}}}
\newcommand{\reshgl}[2]{\textbf{{#1}} \textcolor{green}{($+{#2}$)}}
\newcommand{\reshr}[2]{{#1} \tiny{\textcolor{red}{($-{#2}$)}}}
\newcommand{\rownumber}[1]{\textbf{\textcolor{rowcolor}{#1}}}

\subsection{Implementation details and evaluation setup}
\label{subsec: setup}

\noindent\textbf{Implementation details:} We evaluate our approach on top of two backbones -- VGG-16~\cite{simonyan2013deep} for fair comparison with~\cite{goyal2019counterfactual} and ResNet-50~\cite{he2016deep} for generalizability. As mentioned in Sec.~\ref{subsec: method_preliminaries}, we split both networks into components $f$ and $g$ after the final down-sampling layer \texttt{max\_pooling2d\_5} in VGG-16 and at \texttt{conv5\_1} in ResNet-50. The input images are of size $224 \times 224$ pixels and the output features of $f$ have spatial dimensions $7 \times 7$. We examine counterfactual examples for query-distractor class pairs obtained via the confusion matrix -- for a query class $c$, we select the distractor class $c'$ as the class with which images from $c$ are most often confused. This procedure differs from the approach in~\cite{goyal2019counterfactual} which uses attribute annotations to select the classes $c, c'$. Our setup is more generic as it does not use extra annotations. Distractor images are picked randomly from $c'$.

\noindent \textbf{Auxiliary model:} We adopt the pre-trained ResNet-50~\cite{he2016deep} model from DeepCluster~\cite{caron2020unsupervised} to measure the semantic similarity of the spatial cells. We remove the final pooling layer and apply up- or down-sampling to match the $7 \times 7$ spatial dimensions of features from $f$. As in~\cite{caron2020unsupervised}, we use $\tau=0.1$ in the non-parametric softmax (Eq.~\ref{eq: similarity_loss}). The weight $\lambda=0.4$ (Eq.~\ref{eq: total_loss}) is found through grid search. We set $k = 10$ and select top-10\% most similar cell pairs to pre-filter.

\noindent{\textbf{Evaluation metrics:}} We follow the evaluation procedure from~\cite{goyal2019counterfactual} and report the following metrics using keypoint part annotations.
\begin{itemize}
 \item{\textbf{Near-KP:}} measures if the image regions contain object keypoints (KP). This is a proxy for how often we select discriminative cells, i.e., spatial cells that can explain the class differences.
 \item{\textbf{Same-KP:}} measures how often we select the same keypoints in the query and distractor image, thus measures semantic consistency of counterfactuals. 
 \item{\textbf{\#Edits:}} the average number of edits until the classification model predicts the distractor class $c'$ on the transformed image $I^*$.
\end{itemize}

\begin{wraptable}{l}{0.51\textwidth}
  \begin{center}
    \setlength\tabcolsep{1pt}
    \vspace{-3.5em}
    \caption{\footnotesize{\textbf{Datasets overview.}}}
    \vspace{-1.0em}
    \label{tab: datasets}
    \centering
    \begin{tabular}{l cc lclcl c ccc@{}}
    \toprule
    \textbf{Dataset} &&& \multicolumn{5}{c}{\textbf{Statistics}} && \multicolumn{3}{c}{\textbf{Top-1}} \\
    \cmidrule{3-8} \cmidrule{10-12}
    &&& \tiny{\#Class} && \tiny{\#Train} && \tiny{\#Val} && \tiny{VGG-16} && \tiny{Res-50} \\
    \midrule
    CUB &&& 200 && 5,994 && 5,794 && 81.5 && 82.0 \\
    iNat. (Birds) &&& 1,486 && 414 k && 14,860 && 78.6 && 78.8 \\
    Stanf. Dogs &&& 120 && 12 k && 8,580 && 86.7 && 88.4 \\
    \bottomrule
    \end{tabular}
    \vspace{-3.0em}
  \end{center}
\end{wraptable}									

\noindent\textbf{Datasets:} We evaluate the counterfactuals on three datasets for fine-grained image classification (see Table~\ref{tab: datasets}). The CUB dataset~\cite{CUB} consists of images of 200 bird classes. All images are annotated with keypoint locations of 15 bird parts. The iNaturalist-2021 birds dataset~\cite{INATURALIST} contains 1,486 bird classes and more challenging scenes compared to CUB, but lacks keypoint annotations. So we hired raters to annotate bird keypoint locations for 2,060 random val images from iNaturalist-2021 birds and evaluate on this subset. Stanford Dogs~\cite{DOGS} contains images of dogs annotated with keypoint locations~\cite{DOGSEXTRA} of 24 parts. The explanations are computed on the validation splits of these datasets. 

\subsection{State-of-the-art comparison}
\label{subsec: sota}
\begin{figure}[t]
\centering
\includegraphics[width=0.98\textwidth]{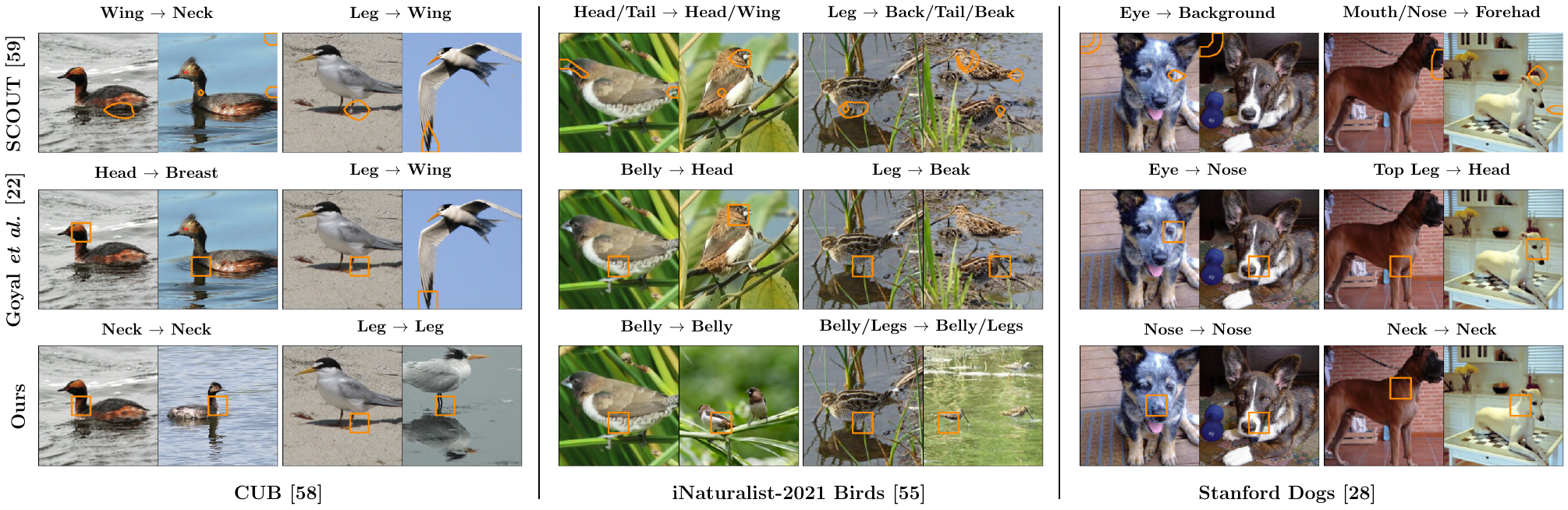}
\vspace*{-1.0em}
\caption{\footnotesize{\textbf{State-of-the-art comparison} of counterfactual explanations (Single Edit - VGG-16). Part labels are included only for better visualization. Image credit:~\cite{inatimages_license}}}
\vspace*{-1.0em}
\label{fig: sota}
\end{figure}

Table~\ref{table: sota} compares our method to other competing methods. We report the results for both (i) the \textbf{single edit} found by solving Eq.~\ref{eq: total_loss} once and (ii) \textbf{all edits} found by repeatedly solving Eq.~\ref{eq: total_loss} until the model's decision changes. Our results are directly comparable with~\cite{goyal2019counterfactual}. By contrast, SCOUT~\cite{wang2020scout} returns heatmaps that require post-processing. We follow the post-processing from~\cite{wang2020scout} where from the heatmaps, select those regions in $I$ and $I'$ that match the area of a single cell edit to compute the metrics. From Table~\ref{table: sota}, we observe that our method consistently outperforms prior works across all metrics and datasets. As an example, consider the \textbf{all edits} rows for the CUB dataset in Table~\ref{table: sota_vgg}. The Near-KP metric improved by $\mathbf{13.9\%}$ over~\cite{goyal2019counterfactual}, indicating that our explanations select more discriminative image regions. More importantly, the Same-KP metric improved by $\mathbf{27\%}$ compared to~\cite{goyal2019counterfactual}, demonstrating that our explanations are significantly more semantically consistent. The average number of edits have also reduced from $5.5$ in~\cite{goyal2019counterfactual} to $\mathbf{3.9}$, meaning that our explanations require fewer changes to $I$ and are thus sparser, which is a desirable property of counterfactuals~\cite{miller2019explanation,wachter2018counterfactual}. Similar performance trends hold on the other two datasets and architectures (Table~\ref{table: sota_resnet}) indicating the generalizability of the proposed approach. Figure~\ref{fig: sota} shows a few qualitative examples where we note that our method consistently identifies semantically matched and class-specific image regions, while explanations from~\cite{goyal2019counterfactual}~and~\cite{wang2020scout} often select regions belonging to different parts.

\setlength{\tabcolsep}{2pt}
\begin{table*}[t]
\centering
\caption{\footnotesize{\textbf{State-of-the-art comparison} against our full proposed pipeline.}}
\vspace*{-1.0em}
\label{table: sota}
\begin{subtable}[t]{0.9\linewidth}
\caption{\footnotesize{Comparison of visual counterfactuals using a VGG-16 model.}}
\label{table: sota_vgg}
\vspace*{-.7em}
\resizebox{\columnwidth}{!}{
\begin{tabular}{c l c ccc c ccc c ccc @{}}
\toprule
& \multirow{2}{*}{\textbf{Method}} && \multicolumn{3}{c}{\textbf{CUB-200-2011}} && \multicolumn{3}{c}{\textbf{INaturalist-2021 Birds}} && \multicolumn{3}{c}{\textbf{Stanford Dogs Extra}}\\
\cmidrule{4-6} \cmidrule{8-10} \cmidrule{12-14}
& && Near-KP & Same-KP & \# Edits && Near-KP & Same-KP & \# Edits && Near-KP & Same-KP & \# Edits\\
\midrule
\noalign{\smallskip}
\multirow{3}{*}{\shortstack[l]{Single\\Edit}} & SCOUT~\cite{wang2020scout} && 68.1 & 18.1 & - && 74.3 & 23.1 & - && 41.7 & 5.5 & - \\
& Goyal~\etal~\cite{goyal2019counterfactual} && 67.8 & 17.2 & - && 78.3 & 29.4 & - && 42.6 & 6.8 & -\\
& Ours && \textbf{73.5} & \textbf{39.6} & - && \textbf{83.6} & \textbf{51.0} & - && \textbf{49.8} & \textbf{23.5} & -\\
\midrule 
\multirow{2}{*}{\shortstack[l]{All\\Edits}} & Goyal~\etal~\cite{goyal2019counterfactual} && 54.6 & 8.3 & 5.5 && 55.2 & 11.5 & 5.5 && 35.7 & 3.7 &\textbf{ 6.3} \\
& Ours && \textbf{68.5} & \textbf{35.3} & \textbf{3.9} && \textbf{70.4} & \textbf{36.9} & \textbf{4.3} && \textbf{37.5} & \textbf{16.4} & 6.6 \\
\bottomrule
\end{tabular}}
\end{subtable}

\vspace*{-0.8em}

\begin{subtable}[t]{0.9\linewidth}
\caption{\footnotesize{Comparison of visual counterfactuals using a ResNet-50 model.}}
\label{table: sota_resnet}
\vspace*{-.7em}
\resizebox{\columnwidth}{!}{
\begin{tabular}{c l c ccc c ccc c ccc @{}}
\toprule
& \multirow{2}{*}{\textbf{Method}} && \multicolumn{3}{c}{\textbf{CUB-200-2011}} && \multicolumn{3}{c}{\textbf{INaturalist-2021 Birds}} && \multicolumn{3}{c}{\textbf{Stanford Dogs Extra}}\\
\cmidrule{4-6} \cmidrule{8-10} \cmidrule{12-14}
& && Near-KP & Same-KP & \# Edits && Near-KP & Same-KP & \# Edits && Near-KP & Same-KP & \# Edits\\
\midrule
\noalign{\smallskip}
\multirow{3}{*}{\shortstack[l]{Single\\Edit}}& SCOUT~\cite{wang2020scout} && 43.0 & 4.4 & - && 53.9 & 8.8 & - && 35.3 & 3.1 & - \\
& Goyal~\etal~\cite{goyal2019counterfactual} && 61.4 & 11.5 & - && 70.5 & 17.1 & - && 42.7 & 6.4 & -\\
& Ours && \textbf{71.7} & \textbf{36.1} & - && \textbf{79.2} & \textbf{33.3} & - && \textbf{51.2} & \textbf{22.6} & -\\
\midrule 
\multirow{2}{*}{\shortstack[l]{All\\Edits}} & Goyal~\etal~\cite{goyal2019counterfactual} && 50.9 & 6.8 & 3.6 && 56.3 & 10.4 & 3.3 && 34.9 & 3.6 & \textbf{4.3} \\
& Ours && \textbf{60.3} & \textbf{30.2} & \textbf{3.2} && \textbf{70.9} & \textbf{32.1} & \textbf{2.6} && \textbf{37.2} & \textbf{16.7} & 4.8\\
\bottomrule
\end{tabular}}
\end{subtable}
\vspace{-0.5em}
\end{table*}

\subsection{Ablation studies}
\label{subsec: ablation}
We now study the different design choices of our framework with~\cite{goyal2019counterfactual} as our baseline and use a VGG-16 model for consistent evaluation on CUB.

\noindent \textbf{Analysis of different components.} Table~\ref{tab: components} reports different variants as we add or remove the following components: semantic loss (Sec.~\ref{subsec: method_consistency}), multiple distractor images (Sec.~\ref{subsec: method_multi_distractor}), and pre-filtering cells (Sec.~\ref{subsec: method_multi_distractor}). Our baseline~\cite{goyal2019counterfactual} (row~\rownumber{1}) establishes a performance limit for the Near-KP and number of edits under the single-distractor setup as the image regions are selected solely based on the predicted class probabilities $g_{c'}(\cdot)$. First, we observe that the semantic loss improves the semantic meaningfulness of the replacements (row~\rownumber{2}), i.e., the Same-KP metric increases by $\textbf{13.7\%}$. However, the Near-KP slightly decreases by $2.5\%$ and the number of edits increases by $1.3$. This may be due to the fact that row~\rownumber{2} considers both the class probabilities $g_{c'}$ and semantic consistency, thereby potentially favoring semantically similar cells over dissimilar cells that yield a larger increase in $g_{c'}$. Second, from rows~\rownumber{1} and \rownumber{3}, we find that allowing multiple distractor images enlarges the search space when solving Eq.~\ref{eq: total_loss}, resulting in better solutions that are more discriminative (Near-KP $\uparrow$), more semantically consistent (Same-KP $\uparrow$) and sparser (fewer edits). Combining the semantic loss with multiple distractor images (row \rownumber{4}) further boosts the metrics. However, using multiple distractor images comes at a significant increase in runtime (almost by 10X). We address this by filtering out semantically dissimilar cell pairs. Indeed, comparing rows~\rownumber{4} and~\rownumber{5}, we note that the runtime improves significantly while maintaining the performance. Putting everything together, our method outperforms~\cite{goyal2019counterfactual} across all metrics (row \rownumber{1} vs. row \rownumber{5}) and generates explanations that are sparser, more discriminative, and more semantically consistent.

\setlength{\tabcolsep}{2pt}
\begin{table}[t]
 \begin{center}
 \caption{\footnotesize{\textbf{Effect of different components of our method:} Row \rownumber{1} is our baseline from~\cite{goyal2019counterfactual}. Our method (row \rownumber{5}) uses multiple distractor images combined pre-filtering irrelevant cells and semantic consistency loss. Time measured on a single V-100 GPU.}}
 \label{tab: components}
 \begin{tabular}{l ccc c llll@{}}
 \toprule
 \textbf{Row \# } & \multirow{2}{*}{\shortstack[c]{\textbf{Semantic} \\ \textbf{Loss}}} & \multirow{2}{*}{\shortstack[c]{\textbf{Multi} \\ \textbf{Distractor}}} & \multirow{2}{*}{\shortstack[c]{\textbf{Filters}\\ \textbf{Cells}}}&& \textbf{Near-KP} & \textbf{Same-KP} & \textbf{Time} (s) & \textbf{\#Edits} \\
 \\
 \midrule
 \rownumber{1} & \xmark & \xmark & \xmark && 54.6 & 8.3 & 0.81 & 5.5 \\
 \rownumber{2} & \cmark & \xmark & \xmark && \reshr{52.1}{2.5} & \reshg{22.0}{13.7} & 1.02 & 6.8 \\
 \rownumber{3} & \xmark & \cmark & \xmark && \reshg{65.6}{11.0} & \reshg{13.8}{5.5} & 9.98 & 3.5 \\
 \rownumber{4} & \cmark & \cmark & \xmark && \reshg{69.2}{14.6} & \reshg{36.0}{23.7} & 10.82  & 3.8\\
 \rownumber{5} & \cmark & \cmark & \cmark && \reshg{68.5}{13.9} & \reshg{35.3}{23.0} & 1.15 & 3.9 \\
 \bottomrule
 \end{tabular}
 \end{center}
\end{table}

\noindent \textbf{Auxiliary model:} Recall from Sec.~\ref{subsec: method_consistency} that representations from self-supervised models efficiently capture richer semantic similarity between local image regions compared to those from supervised models. We empirically verify this by using different pre-training tasks to instantiate the auxiliary model: (i) supervised pre-training with class labels, (ii) self-supervised (SSL) pre-training~\cite{caron2018deep,caron2020unsupervised,he2020momentum} with no labels, and (iii) supervised parts detection with keypoint annotations. We train the parts detector to predict keypoint presence in the $h\times w$ spatial cell matrix using keypoint annotations. We stress that the parts detector is used only as an \textit{upperbound} as it uses part ground-truth to model the semantic constraint. 

We evaluate each auxiliary model by: (i) measuring the Same-KP metric to study if this model improves the semantic matching, and (ii) measuring clustering accuracy to capture the extent of semantic part disentanglement. To measure the clustering accuracy, we first cluster the $d$-dimensional cells in a $7 \times 7$ spatial matrix from $u(\cdot)$ of all images via K-Means and assign each spatial cell to a cluster. Then, we apply majority voting and associate each cluster with a semantic part using the keypoint annotations. The clustering accuracy measures how often the cells contain the assigned part. From Table~\ref{tab: auxiliary_model}, we observe that better part disentanglement (high clustering accuracy) correlates with improved semantic matching in the counterfactuals (high Same-KP). Thus, embeddings that disentangle parts are better suited for the semantic consistency constraint via the non-parametric softmax in Eq.~\ref{eq: similarity_loss}. The CUB classifier fails to model our constraint because it distinguishes between different types of beaks, wings, etc., to optimize for the classification task (Same-KP drops by 12.1\% vs. the upperbound). Differently, the SSL features are more generic, making them suitable for our method (Same-KP using DeepCluster drops only 0.2\% vs. the upperbound).

\setlength{\tabcolsep}{4pt}
\begin{table}[t]
 \caption{\footnotesize{\textbf{Comparison of auxiliary models on CUB:} We study the Same-KP metric of the counterfactuals (single distractor) and whether the aux. features can be clustered into parts. $^\dagger$Parts detector establishes an upperbound as it uses parts ground-truth.}}
 \label{tab: auxiliary_model}
 \centering
 \begin{tabular}{lc lc cc ccccc}
 \toprule
 \textbf{Auxiliary Model} && \textbf{Annotations} && \textbf{Counterfactuals} && \multicolumn{5}{c}{\textbf{Clustering} (K-Means Acc.)} \\
 \cmidrule{5-5} \cmidrule{7-11}
 && && (Same-KP) && K=15 && K=50 && K=250 \\
 \midrule
 CUB Classifier && Class labels && 10.1 && 18.0 && 19.3 && 21.6 \\
 IN-1k Classifier && Class labels && 19.3 && 42.0 && 49.5 && 57.1 \\
 \midrule
 IN-1k MoCo~\cite{he2020momentum} && None (SSL) && 18.1 && 33.8 && 44.1 && 52.2 \\
 IN-1k SWAV~\cite{caron2020unsupervised} && None (SSL) && 22.1 && 45.3 && 54.2 && 62.6 \\
 IN-1k DeepCluster~\cite{caron2018deep} && None (SSL) && 22.0 && 45.3 && 54.9 && 63.5 \\
 \midrule
 CUB Parts Detector$^\dagger$ && Keypoints && 22.2 && 46.0 && 59.2 && 75.4 \\
 \bottomrule
 \end{tabular}
\end{table}

\begin{wrapfigure}{l}{0.56\textwidth}
  \begin{center}
    \vspace{-3.2em}
    \begin{subfigure}[b]{0.27\textwidth}
    \centering
    \includegraphics[width=\textwidth]{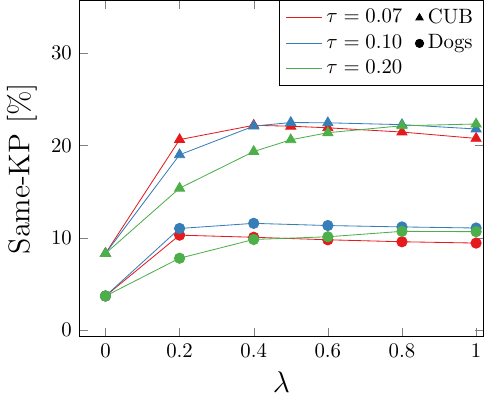}
    \end{subfigure}
    \hfill
    \begin{subfigure}[b]{0.27\textwidth}
    \centering
    \includegraphics[width=\textwidth]{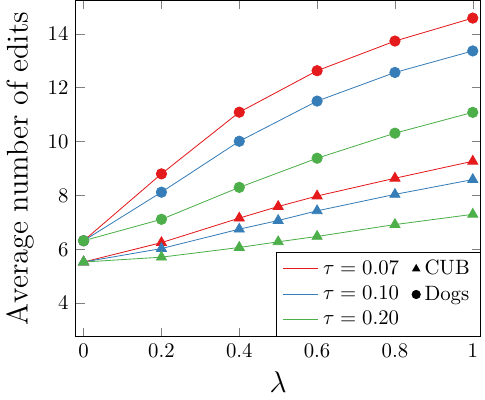}
    \end{subfigure}
    \vspace{-1.2em}
    \caption{\footnotesize{Influence of temperature $\tau$ and weight $\lambda$.}}
    \label{fig: soft_constraint_ablation}
    \vspace{-2.8em}
  \end{center}
\end{wrapfigure}

\noindent \textbf{Influence of $\mathbf{\tau}$ and $\mathbf{\lambda}$:}
We study how the temperature $\tau$ in Eq.~\ref{eq: similarity_loss} and the weight $\lambda$ parameter in Eq.~\ref{eq: total_loss} influence different metrics. Recall that high values of $\lambda$ favor the semantic loss over the classification loss. Selecting semantically similar cells over dissimilar ones directly improves the Same-KP metric (Fig.~\ref{fig: soft_constraint_ablation} (left)), but that comes at a cost of an increased number of edits until the model's decision changes (Fig.~\ref{fig: soft_constraint_ablation} (right)). We observe that $\lambda=0.4$ is a saturation point, after which the Same-KP metric does not notably change. Further, lower values of $\tau$ sharpen the softmax distribution making it closer to one-hot, while higher $\tau$ yield a distribution closer to a uniform. This has an effect on the number of edits, as a sharper distribution is more selective. We found that for a fixed $\lambda=0.4$, $\tau=0.1$ as in~\cite{caron2020unsupervised} is a sweet spot between good Same-KP performance and a small increase in the number of edits. We verified values via 5-fold cross-validation across multiple datasets.

\subsection{Online evaluation through machine teaching}
\label{subsec: machine_teaching}
To further demonstrate the utility of high-quality visual counterfactuals, we setup a machine teaching experiment, where humans learn to discern between bird species with the help of counterfactual explanations.
Through the experiment detailed below, we verify our hypothesis that humans perform better at this task with more informative and accurate counterfactual explanations.

\begin{figure}[t]
\begin{minipage}[t]{.48\linewidth} 
    \centering
    \includegraphics[width=\textwidth]{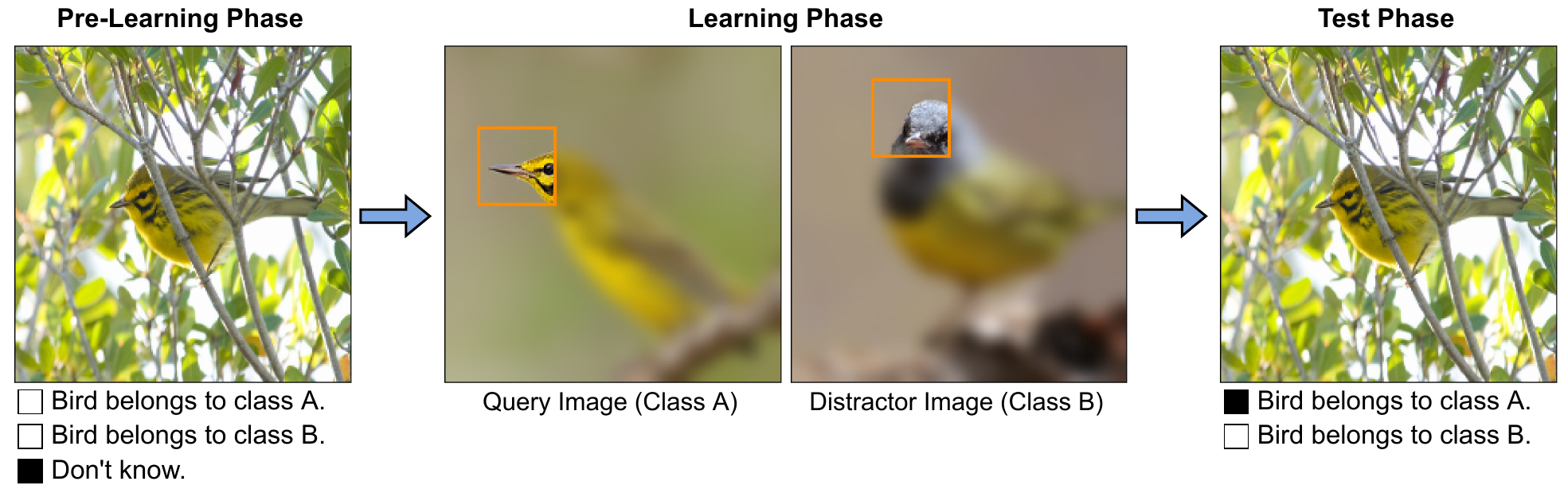}
    \vspace{-2.3em}
    \caption{\footnotesize{Machine teaching task phases.}}
    \label{fig: machine_teaching}
\end{minipage} %
\hfill
\begin{minipage}[t]{.48\linewidth} 
    \centering
    \includegraphics[width=\textwidth]{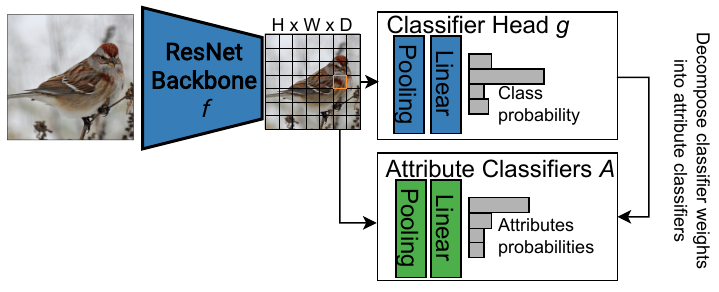}
    \vspace{-2.3em}
    \caption{\footnotesize{Attribute-based decomposition.}}
    \label{fig: attributes_method}
\end{minipage} %
\end{figure}

\noindent\textbf{Study setup:}
We follow the setup from~\cite{wang2020scout}, but differ in two crucial ways: (i) ours is a larger study on 155 query-distractor class pairs, while~\cite{wang2020scout} was done only on one class pair; (ii) we obfuscate the bird class names and replace them with ``class A'' and ``class B''. We do this because some class names contain identifiable descriptions (e.g., \emph{Red Headed Woodpecker}) without needing visual cues. 
The study comprises three phases (simplified visualization in Fig.~\ref{fig: machine_teaching}). The \textbf{pre-learning phase} gives AMT raters 10 \underline{test} image examples of 2 bird classes. The raters need to choose one of three options: `Bird belongs to class A', `Bird belongs to class B,' or `Don't know'. The purpose of this stage is for the raters to get familiarized with the user interface, and as in~\cite{wang2020scout} all raters chose `Don’t know' for each example in this stage. Next, during the \textbf{learning phase}, we show counterfactual explanations of 10 \underline{train} image pairs where the query image belongs to class A and the distractor image to class B. 
We highlight the image content from the counterfactual region, with all other content being blurred (Fig.~\ref{fig: machine_teaching}). This ensures that the humans do not perform the classification task based on any other visual cues except the ones identified by a given counterfactual method. Finally, the \textbf{test phase} presents to raters 10 \underline{test} image pairs (same as in the pre-learning stage), and asks to classify them into either class A or B. This time, the option `Don't know' is not provided. Once the task is done, a different set of bird class pair is selected, and the three stages are repeated. 

\begin{table}[b]
    \vspace{-2.0em}
    \begin{minipage}[t]{0.56\linewidth}
        \centering
        \caption{\footnotesize{\textbf{Machine teaching task.} The learning phase selects random image pairs ($\dagger$), or pairs that show the largest improvement in terms of being semantically consistent ($*$).}}
        \label{tab: machine_teaching}
        \begin{tabular}{l c cc@{}}
        \toprule
        \textbf{Method} && \multicolumn{2}{c}{\textbf{Test Acc.} (\%)}\\
        && (Random)$^\dagger$ & (Semantically-acc.)$^*$ \\
        \midrule
        SCOUT~\cite{wang2020scout} && 77.4 & 62.8 \\
        Goyal~\etal~\cite{goyal2019counterfactual} && 76.7 & 64.3 \\
        Ours  && \textbf{80.5} & \textbf{82.1} \\
        \bottomrule
        \end{tabular}
    \end{minipage}
    \hfill
    \begin{minipage}[t]{0.40\linewidth}
        \centering
        \caption{\footnotesize{\textbf{Attribute-based counterfactuals.} We evaluate whether the top-1 attributes are discriminative of the classes.}}
        \label{tab: attributes}
        \vspace{0.9em}
        \begin{tabular}{l c c @{}}
        \toprule
        \textbf{Method} && \textbf{Test Acc.} (\%) \\
        \midrule
        SCOUT~\cite{wang2020scout} && 46.7 \\
        Goyal~\etal~\cite{goyal2019counterfactual}   && 67.0 \\
        Ours    && \textbf{74.5} \\
        \bottomrule
        \end{tabular}
    \end{minipage}
\end{table}

\noindent\textbf{Task details:} We hired 25 AMT raters, use images from CUB, and compare counterfactuals produced from our method with two baselines:~\cite{goyal2019counterfactual} and~\cite{wang2020scout}. For all three methods, we mine query-distractor classes via the approach mentioned in Sec.~\ref{subsec: setup}, resulting in 155 unique binary classification tasks. The learning phase visualizes the counterfactual generated from the first edit. To ensure a fair comparison across all methods, we do not use multiple distractor images for generating counterfactuals, use the exact same set of images across all the compared methods, and use the same backbone (VGG-16~\cite{simonyan2014very}) throughout. This controlled setup ensures that any difference in the human study performance can be only due to the underlying counterfactual method. We report results under two setups, which differ in how we select the image pairs $(I, I')$: \textbf{1. random}: we generate explanations from random images using different methods. This is a fair comparison between all methods. \textbf{2. semantically-consistent}: we study whether semantically consistent explanations lead to better human teaching. Hence, we exaggerate the differences in Same-KP between our method and~\cite{goyal2019counterfactual,wang2020scout} by selecting images where our approach has a higher Same-KP metric. If semantic consistency is important in machine teaching, our approach should do much better than `random', and the baselines should do worse than `random'.


\noindent\textbf{Results:} Table~\ref{tab: machine_teaching} shows that the raters perform better when shown explanations from our method under the `random' setup. Further, the differences in test accuracy are more pronounced ($82.1\%$ vs. $64.3\%$) when the raters were presented with semantically consistent explanations. This result highlights the importance of semantically consistent visual counterfactuals for teaching humans.


\section{Towards language-based counterfactual explanations}
\label{sec: attributes}
In this section, we propose a novel method to augment visual counterfactual explanations with natural language via the vocabulary of parts and attributes. Parts and attributes bring notable benefits as they enrich the explanations and make them more interpretable~\cite{kim2018interpretability}. Through this experiment, we further emphasize the importance of semantically consistent counterfactuals and prove them to be a key ingredient towards generating natural-language-based explanations.

Our proof-of-concept experiment uses a ResNet-50 model, where $f(\cdot)$ computes the $h\times w\times d$ spatial feature output of the last conv layer, and $g(\cdot)$ performs a global average pooling operation followed by a linear classifier. We use the CUB~\cite{CUB} dataset with 15 bird parts, where each part (e.g., beak, wing, belly, etc.) is associated with a keypoint location. Additionally, this dataset contains part-attribute annotations (e.g., hooked beak, striped wing, yellow belly, etc.). We perform our analysis on a subset of 77 subsequently denoised part-attributes. Following~\cite{koh2020concept}, denoising is performed by majority voting, e.g., if more than 50\% of crows have black wings in the data, then all crows are set to have black wings.

In the first step, given a query $I$ from class $c$ and a distractor $I'$ from $c'$, we construct a counterfactual $I^*$, following our approach from Sec.~\ref{sec: method}. For fair comparison with~\cite{goyal2019counterfactual,wang2020scout}, we limit to single best cell edits. Next, we identify the part corresponding to this best-edit cell in $I$. We train a parts detector that predicts the top-3 parts for each cell location in the $h\times w$ spatial grid. Note that if the corresponding cell in $I'$ is not semantically consistent with $I$, the detected parts will not match, and the attribute explanations will be nonsensical. Finally, we find \emph{the most important} attribute for the best-edit via the procedure below.

{\noindent{\textbf{Finding the best attribute:}}} 
We train a part-attribute model $A$ that performs global average pooling followed on the output of $f(.)$ by a linear classifier, thus operating on the same feature space as $g$. We then use an interpretable basis decomposition~\cite{zhou2016learning} to decompose the object classifier weights from $g(\cdot)$ into a weighted sum of part-attribute classifier weights from $A(\cdot)$.
A simplified visualization is presented in Fig.~\ref{fig: attributes_method}, see~\cite{zhou2016learning} for details. 
The interpretable basis decomposition yields an importance score $s_t$ for each part-attribute $t$, and we additionally constrain the part-attributes to only the detected parts in the best-edit cells. E.g., if the detected part is a beak, we only consider the \{hooked, long, orange, ...\}-beak attribute classifiers. Similarly, we compute an importance score $s'_t$ for the best-edit cell in $I'$. Finally, we compute the differences of importance scores $s'_t - s_t$, where a positive difference indicates that part-attribute $t$ contributed more to the model's decision being $c'$ compared to $c$. We select the top-$k$ such part-attributes. Again, note that the difference $s'_t - s_t$ makes sense only if the selected parts are semantically same in $I$ and $I'$ (details in suppl.).

\noindent\textbf{Evaluation:} 
For each class pair $(c, c')$, we use the available annotations to define part-attributes that belong to class $c$ but not to class $c'$, and vice-versa, as proxy counterfactual ground-truth. Our final explanations are evaluated by measuring how often the top-1 part-attribute, identified via the difference between the estimated importance scores, belongs to the set of ground-truth part-attributes.

\begin{figure}[t]
    \centering
    \includegraphics[width=0.82\linewidth]{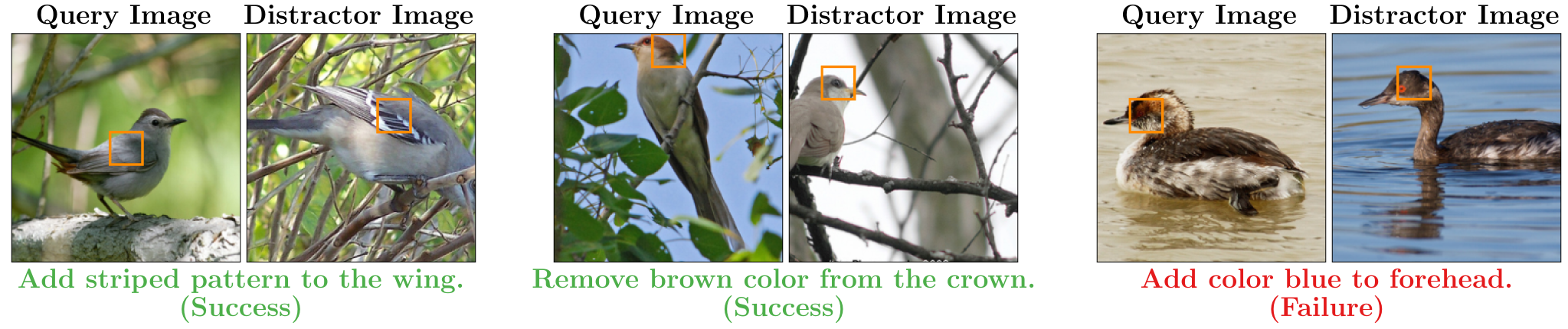}
    \vspace{-1.2em}
    \caption{\footnotesize{\textbf{Augmenting counterfactuals with part-attributes.} We identify the attribute that is most important for changing the model's decision. Best viewed in color.}}
    \label{fig: attributes}
    \vspace{-1.0em}
\end{figure}

\noindent\textbf{Results:} Table~\ref{tab: attributes} shows the results using visual counterfactuals from our method and from~\cite{goyal2019counterfactual,wang2020scout}. We observe that our method is significantly better compared to prior work in correctly identifying discriminative part-attributes. Given that all other factors were controlled across the three methods, we argue that this improvement is due to our counterfactuals being semantically consistent. Figure~\ref{fig: attributes} shows the qualitative results. Notice that both the wing's color and pattern are visually distinct in Fig.~\ref{fig: attributes} (left), but the part-attribute explanation points out that the wing's pattern mattered the most to the model while generating the counterfactual. Similarly in~Fig.~\ref{fig: attributes} (middle), the part-attribute explanation tells us that the crown color is most important.
In both cases, the part-attribute information helps disambiguate the visual explanation. Figure~\ref{fig: attributes} (right) shows a failure case caused by a wrongful prediction from the part-attribute classifiers.

\section{Conclusion and future work}
\label{sec: conclusion}
We presented a novel framework to generate semantically consistent visual counterfactuals. Our evaluation shows that (i) our counterfactuals consistently match semantically similar and class-specific regions, (ii) our proposed method is computationally efficient, and (iii) our explanations significantly outperform the s-o-t-a. Further, we demonstrated the importance of semantically consistent visual counterfactuals via: (i) a machine teaching task on fine-grained bird recognition, and (ii) an approach to augment our counterfactuals with a human interpretable part and attribute vocabulary. Currently, our method greedily searches for one cell replacement at a time. Relaxing this constraint to explore multiple regions in parallel is a fruitful future research problem. Finally, we only scratched the surface in augmenting visual counterfactuals with attribute information. We hope that our work will spark more interest in this worthy topic by the community.
\clearpage
\bibliographystyle{splncs04}
\bibliography{citations}
\clearpage


\setcounter{section}{0}
\renewcommand\thesection{\Alph{section}}
\setcounter{figure}{0}
\setcounter{table}{0}
\renewcommand{\thefigure}{S\arabic{figure}}
\renewcommand{\thetable}{S\arabic{table}}

\section{Implementation details}
\label{suppl: training}
This section discusses additional implementation details of our experiments in Sec.~\ref{sec: experiments} and Sec.~\ref{sec: attributes}.

\subsubsection{Data.} We use the following publicly available datasets for fine-grained image recognition: CUB~\cite{CUB}, iNaturalist-2021~\cite{INATURALIST}, and StanfordDogs~\cite{DOGS}. The image classifier implementation follows the typical implementation in PyTorch and uses standard image augmentations (i.e., random resized cropping and random horizontal flipping). We provide additional details for each dataset below:
\begin{itemize}
    \item \textbf{CUB}: The CUB dataset consists of images of 200 bird species annotated with keypoint locations of 15 bird parts, e.g., crown, beak, etc. Some of the keypoint annotations distinguish between the left-right instances of parts: `left wing' / `right wing', `left leg' / `right leg', and `left eye' / `right eye'. We treat these as a single part during the evaluation of the Near-KP and Same-KP metrics, i.e., `left wing' and `right wing' as `wing'.
    \item \textbf{iNaturalist-2021}: The iNaturalist-2021 dataset consists of various super-categories (e.g., plants, insects, birds, etc.), covering 10,000 species in total. The dataset contains a larger number of classes and more complex scenes compared to other fine-grained image recognition datasets. Therefore, the iNaturalist-2021 dataset can be considered as a more challenging testbed for our approach. However, this dataset lacks keypoint annotations. We used the bird supercategory for our quantitative evaluation in Sec.~\ref{sec: experiments}, and requested human annotators to provide part keypoint information for 2,060 validation images. The keypoint definitions from the CUB dataset are used. We did not perform a quantitative evaluation of other supercategories, as these are considerably more challenging to annotate. Specifically, we identified the following challenges: (i) some supercategories do not have identifiable parts (e.g., fungi), and (ii) some supercategories are too diverse, and do not support common keypoint definitions across all sub-categories (e.g., plants, mammals, insects, etc.). We do provide qualitative results on several of these supercategories in Sec.~\ref{suppl: qualitative}.
    \item \textbf{StanfordDogs}: The StanfordDogs dataset contains images of 120 dog breeds taken from the ImageNet~\cite{IMAGENET} dataset. The keypoint annotations are provided by~\cite{DOGSEXTRA}. Again, we treat left-right instances of parts as the same part, i.e., left ear and right ear as just ear. 
\end{itemize}

\subsubsection{Classifier.} The training follows the typical VGG-16 and ResNet-50 implementation in PyTorch~\cite{paszke2019pytorch} with 100 epochs. All models use pre-trained ImageNet~\cite{IMAGENET} weights. The training uses stochastic gradient descent with momentum $0.9$ and weight decay of $0.0001$. We use batches of size 32 for the CUB and Stanford Dogs datasets, and batches of size 256 for the iNaturalist-2021 dataset. The initial learning rate is selected via grid search and decreased by 10 at the 70-th and 90-th percentile of training.

\subsubsection{Self-supervised models.} We used the publicly available weights that were provided by the authors of the respective works~\cite{caron2020unsupervised,he2020momentum}. For DeepCluster and SWAV, we adopt the models trained via the multi-crop augmentation from~\cite{caron2020unsupervised}. We pre-trained all models on ImageNet. Different pre-training schemes could be used when considering more specialized domains like medical images.

\subsubsection{Parts detector.} We trained a parts detector on CUB. The parts detector consists of a ResNet-50 backbone followed by a $1 \times 1$ convolutional layer. The input images are $224 \times 224$ pixels and the output has spatial dimensions $7 \times 7$. We project the ground-truth bird keypoint annotations onto a grid of shape $7 \times 7$ and train the parts detector to predict keypoint presence via a multi-class cross-entropy loss. The loss is only applied to cells that contain at least one keypoint. Training follows the classifier implementation but we decrease the number of epochs to 50. The initial learning rate is $0.001$. We evaluate the predictions via the mean AP metric (excluding cells that contain no keypoints). The best model obtains a mean AP of $92.3$ on the validation set.

\section{Additional results}
\label{suppl: additional_results}
We report additional results that complement the ablation studies in the main paper.

\subsubsection{Selection of query-distractor classes.} The experiments in Sec.~\ref{sec: experiments} examine counterfactual examples for query-distractor classes obtained via the confusion matrix - for a query class $c$, we select the distractor class $c'$ as the class with which images from $c$ are most often confused. This procedure differs from the approach in~\cite{goyal2019counterfactual} which uses attribute annotations to select the most confusing classes. In particular, the authors select $c'$ as the nearest neighbor class of $c$ in terms of the average attribute annotations provided with the dataset. We argue that our setup is more generic as it does not use additional annotations. For completeness, we provide results with both selection procedures in Table~\ref{tab: selection}. We observe that there are no significant differences in the results when adjusting the selection procedure. In practice, the two selection procedures often generate the same query-distractor class pairs. In conclusion, our selection procedure provides a viable and more generic alternative to the method from~\cite{goyal2019counterfactual}.

\setlength{\tabcolsep}{2pt}
\begin{table}[t]
 \begin{center}
 \caption{\small{\textbf{Comparison of different methods to select the query-distractor classes.} We select the distractor class as the most confusing class in the confusion matrix, or as the nearest neighbor class in terms of the average attribute annotations. The results are reported for a VGG-16 classifier on CUB.}}
 \label{tab: selection}
 \begin{tabular}{lc lc lll@{}}
 \toprule
 \textbf{Method} && \textbf{Selection Procedure} && \textbf{Near KP} & \textbf{Same KP} & \textbf{\#Edits} \\
 \\
 \midrule
 \multirow{2}{*}{Goyal~\etal~\cite{goyal2019counterfactual}} && Confusion Matrix && 54.6 & 8.3 & 5.5 \\
 && Attributes && 55.0 & 8.6 & 5.4 \\
 \midrule
 \multirow{2}{*}{Ours} && Confusion Matrix && 68.5 & 35.3 & 3.9 \\
 && Attributes && 68.6 & 35.6 & 3.9 \\
 \bottomrule
 \end{tabular}
 \end{center}
 \vspace{-2em}
\end{table}

\subsubsection{Background as a metric.} The CUB dataset contains mask annotations that segment the foreground object. Prior work~\cite{goyal2019counterfactual} used the object segmentation to measure how often the counterfactuals select cells belonging to the foreground object. Like the Same-KP metric, this \emph{foreground metric} is a proxy for how often the counterfactuals select discriminative cells, i.e., cells that explain the class differences. For completeness, we report the results of the foreground metric in Table~\ref{tab: foreground}. Our method outperforms the baseline~\cite{goyal2019counterfactual} in terms of the foreground metric, which indicates that we select more discriminative cells in the image. This observation aligns with our conclusions in the paper based upon the Near-KP metric. Note that the other metrics were discussed in Section~\ref{subsec: ablation}.

\begin{table}[t]
 \begin{center}
 \caption{\small{\textbf{Ablation study of the foreground metric.} The results are with VGG-16 on CUB. We compare the baseline~\cite{goyal2019counterfactual} against our method.}}
 \label{tab: foreground}
 \begin{tabular}{lc llll@{}}
 \toprule
 && \textbf{Foreground} & \textbf{Near KP} & \textbf{Same KP} & \textbf{\#Edits} \\
 \\
 \midrule
 Goyal~\etal~\cite{goyal2019counterfactual} && 94.2 & 54.6 & 8.3 & 5.5 \\
 Ours && \reshg{99.1}{4.9} & \reshg{68.5}{13.9} & \reshg{35.3}{23.0} & 3.9 \\
 \bottomrule
 \end{tabular}
 \end{center}
\end{table}

\subsubsection{Soft versus hard semantic constraint.}
We have modeled the semantic consistency constraint in a \emph{soft} way (see Eq.~\ref{eq: total_loss}). That is, we select replacements that balance the increase of $g_{c'}(\cdot)$ with the semantic similarity of the image regions. Alternatively, the constraint could be implemented in a \emph{hard} way. That is, we cluster the auxiliary spatial features first, e.g., K=50, and only replace query cells with distractor cells from the same cluster. Table~\ref{table: ablation_hard_soft} compares the two mechanisms. The hard constraint selects considerably less discriminative cells (lower Near-KP and more edits) as it's more restrictive of the cells that can be replaced. In contrast, the soft constraint achieves better results, as it balances the $\mathcal{L}_s$ and $\mathcal{L}_c$ losses in Eq.~\ref{eq: total_loss}.

\setlength{\tabcolsep}{2pt}
\begin{table}[ht]
\begin{center}
\caption{\small{\textbf{Comparison of the hard and soft constraint mechanism}. Results are obtained with a VGG-16 classifier on CUB. We use a single distractor image.}}
\label{table: ablation_hard_soft}
\begin{tabular}{l c lllc@{}}
\toprule
\textbf{Constraint} && \textbf{Near KP} & \textbf{Same KP} & \textbf{\#Edits}\\
\midrule
Hard && 40.1 & 13.3 & 9.3 \\
Soft && \textbf{52.1} & \textbf{22.0} & \textbf{6.8} \\
\bottomrule
\end{tabular}
\end{center}
\end{table}

\subsubsection{Clustering.} 
\begin{figure}
    \centering
    \includegraphics[width=\textwidth]{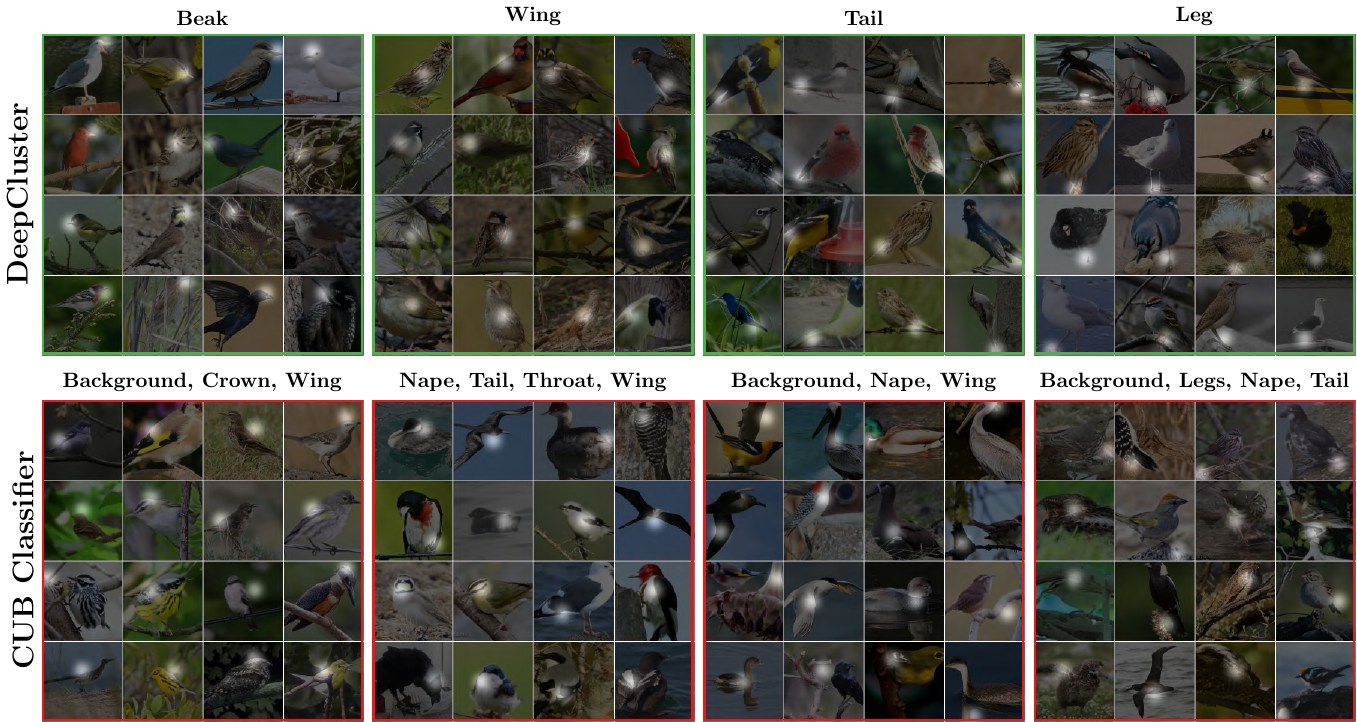}
    \caption{\small{\textbf{Clustering visualizations.} We study part disentanglement when clustering the spatial features obtained with different auxiliary models. We highlight image regions assigned to the same cluster (best viewed in color digitally). We indicate the parts found in each cluster for the purpose of visualization.}}
    \label{fig: cluster_visualizations}
\end{figure}

We studied the part clustering accuracy via different auxiliary models in Table~\ref{tab: auxiliary_model}. In this way, we verified whether the spatial feature representations of the auxiliary models are capable of disentangling parts. In this section, we provide qualitative results of this experiment. Figure~\ref{fig: cluster_visualizations} visualizes clusters found via a CUB classifier and DeepCluster model. We select several clusters and highlight cells assigned to the same cluster. The DeepCluster features better disentangle parts, i.e., cells assigned to the same cluster refer to the same part.

\subsubsection{Ablation of pre-filtering operation.}
We study the influence of the pre-filtering step from Sec.~\ref{subsec: method_multi_distractor} in Table~\ref{table: prefiltering}. Namely, we vary $k$ in the selected top-$k\%$ permutations to be used in the multi-distractor setup. As it can be observed, our method is very robust to the choice of the value $k$, so we selected $k$ in order to achieve around x10 speedup over the vanilla multi-distractor approach.

\setlength{\tabcolsep}{2pt}
\begin{table}[ht]
\begin{center}
\caption{\small{\textbf{Ablation of pre-filtering operation} that selects $k\%$ of permutations via the semantic similarity loss. We study the influence of varying $k\%$.}}
\label{table: prefiltering}
\begin{tabular}{lllll@{}}
\toprule
\textbf{k\%} & \textbf{Near KP} & \textbf{Same KP} & \textbf{\#Edits} & \textbf{Time (s)} \\
\midrule
0.01 & 61.2 & 34.1 & 4.8 & 0.18 \\
0.05 & 66.8 & 34.9 & 4.2 & 0.64 \\
0.10 & 68.5 & 35.3 & 3.9 & 1.15 \\
0.15 & 68.9 & 35.5 & 3.9 & 1.71 \\
0.20 & 69.1 & 35.9 & 3.8 & 2.20 \\
1.00 (no-prefiltering) & 69.2 & 36.0 & 3.8 & 10.82 \\
\bottomrule
\end{tabular}
\end{center}
\end{table}

\subsubsection{Effect of receptive field.} We discuss how the receptive field size of the classification network effects the quality of the counterfactuals. Large receptive fields can lead to poorer localization and reduce the counterfactual's quality. For example, the ResNet-50 models with receptive field size 299 yield lower numbers in Table~\ref{table: sota} compared to their VGG-16 counterparts with receptive field size 212. We tried to mitigate this behavior by using features from the earlier \texttt{conv5\_1} layer instead of \texttt{conv5\_3} for ResNet and found it improves results. For example, on CUB, this change led to a reduction in the number of edits from 8.0 to 3.2. In the future, we could further address this issue by using features from earlier layers with better localization. Alternatively, we could explore techniques which control the receptive field size of the network.

\subsubsection{Interpretation of Same-KP.} Visualization of our counterfactual explanations shows that we consistently identify class-specific and semantically matched parts (see Figure~\ref{fig: sota}). However, the absolute values of the Same-KP metric in Table~\ref{table: sota} might still seem low ($< 40 \%$). There are two main reasons for this. First, we project the keypoint annotations from the query and distractor images onto the discrete spatial cells during evaluation, associating each cell with a set of keypoints. Now, keypoints lying near the cell boundaries are assigned to only one cell. At the time of evaluation, for such a keypoint, even if a neighboring cell is chosen for replacement, Same-KP metric is penalized. Secondly, a keypoint represents just a near center point of a semantic part, rather than the whole part. Hence, a wing of the bird may actually belong to two adjoining spatial cells, but the keypoint is only in one cell. Thus we also report Near-KP metric which does not suffer form these issues. In conclusion, our counterfactuals are faithful, which is also reflected in our qualitative results.

\section{Qualitative results}
\label{suppl: qualitative}
\subsubsection{Additional qualitative examples.} 
\begin{figure}
    \centering
    \includegraphics[width=\textwidth]{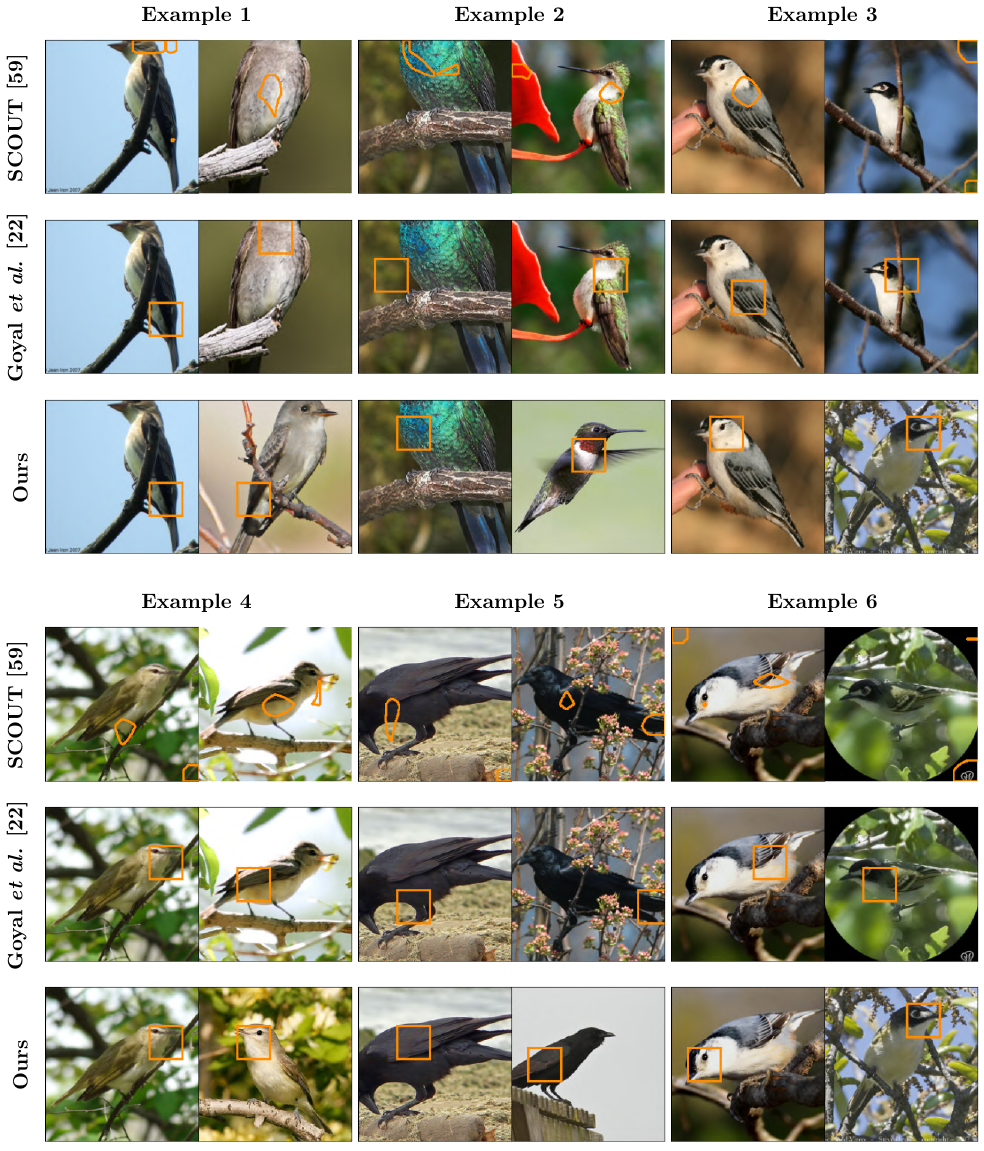}
    \caption{\textbf{Additional qualitative results on CUB~\cite{CUB}.} We highlight the best edit in the query image (left) and distractor image (right).}
    \label{fig: suppl_cub_res50}
\end{figure}

\begin{figure}
    \centering
    \includegraphics[width=\textwidth]{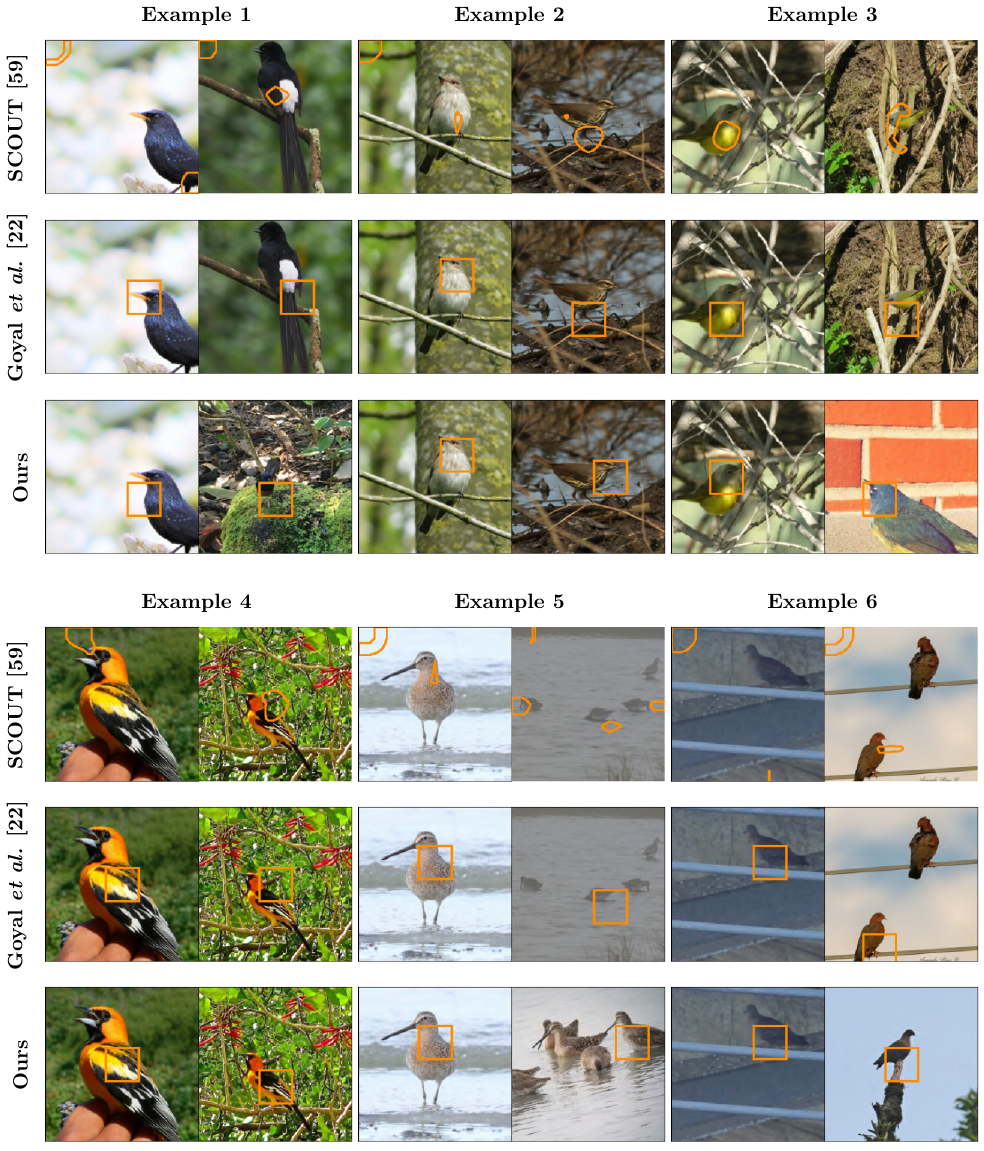}
        \caption{\textbf{Additional qualitative results on iNaturalist-2021 Birds~\cite{INATURALIST}.} We highlight the best edit in the query image (left) and distractor image (right).}
    \label{fig: suppl_inat_birds_res50}
\end{figure}

Figures~\ref{fig: suppl_cub_res50}-\ref{fig: suppl_inat_birds_res50} show counterfactual examples generated with different methods on the CUB~\cite{CUB}, iNaturalist-2021 Birds~\cite{INATURALIST} and StanfordDogs~\cite{DOGS} datasets. The model is ResNet-50. In each case, we highlight the single best edit. We observe that our counterfactual explanations consistently identify class-specific and semantically consistent image regions. This is opposed to other counterfactual explanations~\cite{goyal2019counterfactual,wang2020scout} which often replace regions of different parts. 

\subsubsection{Qualitative examples for other iNaturalist-2021 supercategories.} 
\begin{figure}
    \centering
    \includegraphics[width=\textwidth]{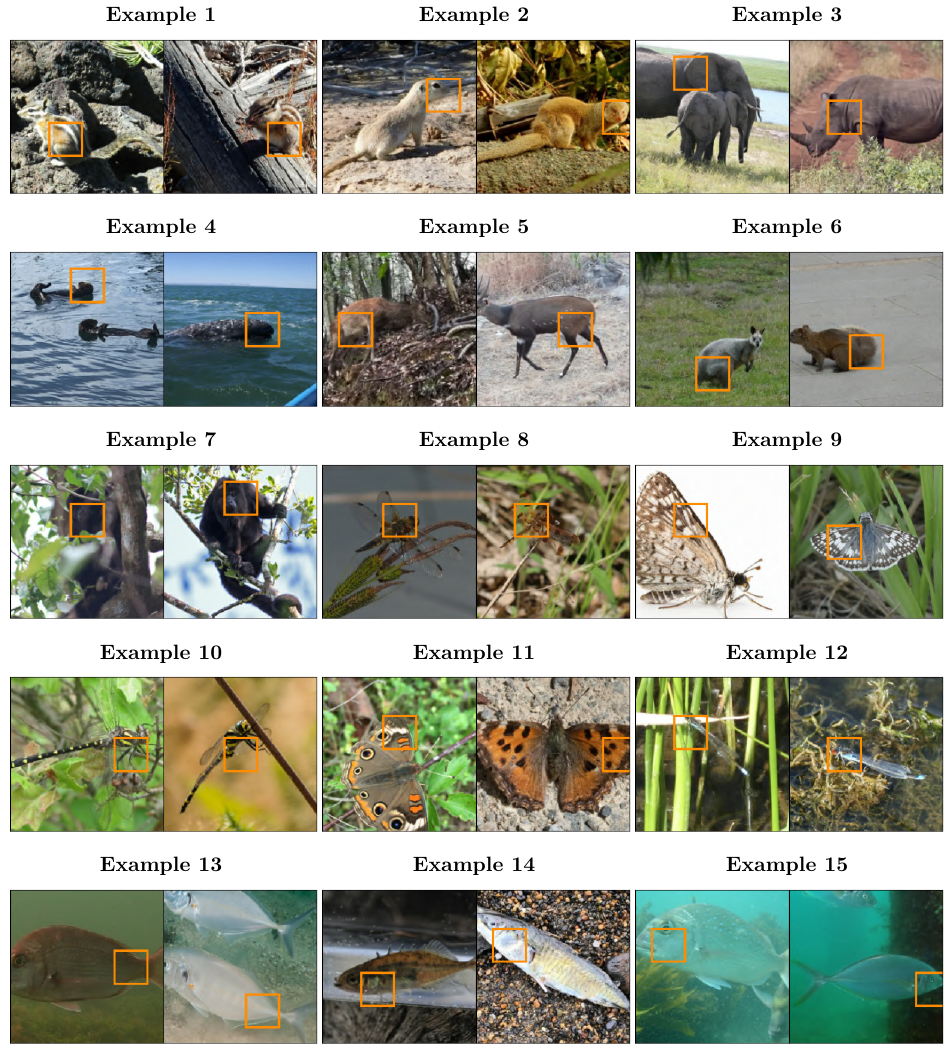}
    \caption{\textbf{Qualitative results on other iNaturalist-2021 supercategories}. We show counterfactual explanations from our method on the following iNaturalist-2021 supercategories: `Mammals', `Ray-finned Fishes' and `Insects'. We highlight the best edit in the query image (left) and distractor image (right).}
    \label{fig: suppl_inat_other_res50}
\end{figure}

The paper only studied the birds supercategory on iNaturalist-2021, as other supercategories are considerably harder to annotate with keypoint information (see discussion in Sec.~\ref{suppl: training}). In this subsection, we demonstrate that our method can be applied to other supercategories too via qualitative results. We train separate image classifiers on the supercategories of `Mammals', `Insects', and `Ray-finned Fishes', and generate counterfactual explanations using our approach. The image classifiers use a ResNet-50 model, and training follows the iNaturalist-2021 implementation detailed in Sec.~\ref{suppl: training}. Figure~\ref{fig: suppl_inat_other_res50} shows the results for the single best edit. Again, we find that our counterfactual explanations highlight class-specific and semantically consistent image regions in the query and distractor images. In conclusion, our method applies to a broad variety of fine-grained image classification tasks.

\subsubsection{Visualization of multiple edits.} 
\begin{figure}
    \centering
    \includegraphics[width=\textwidth]{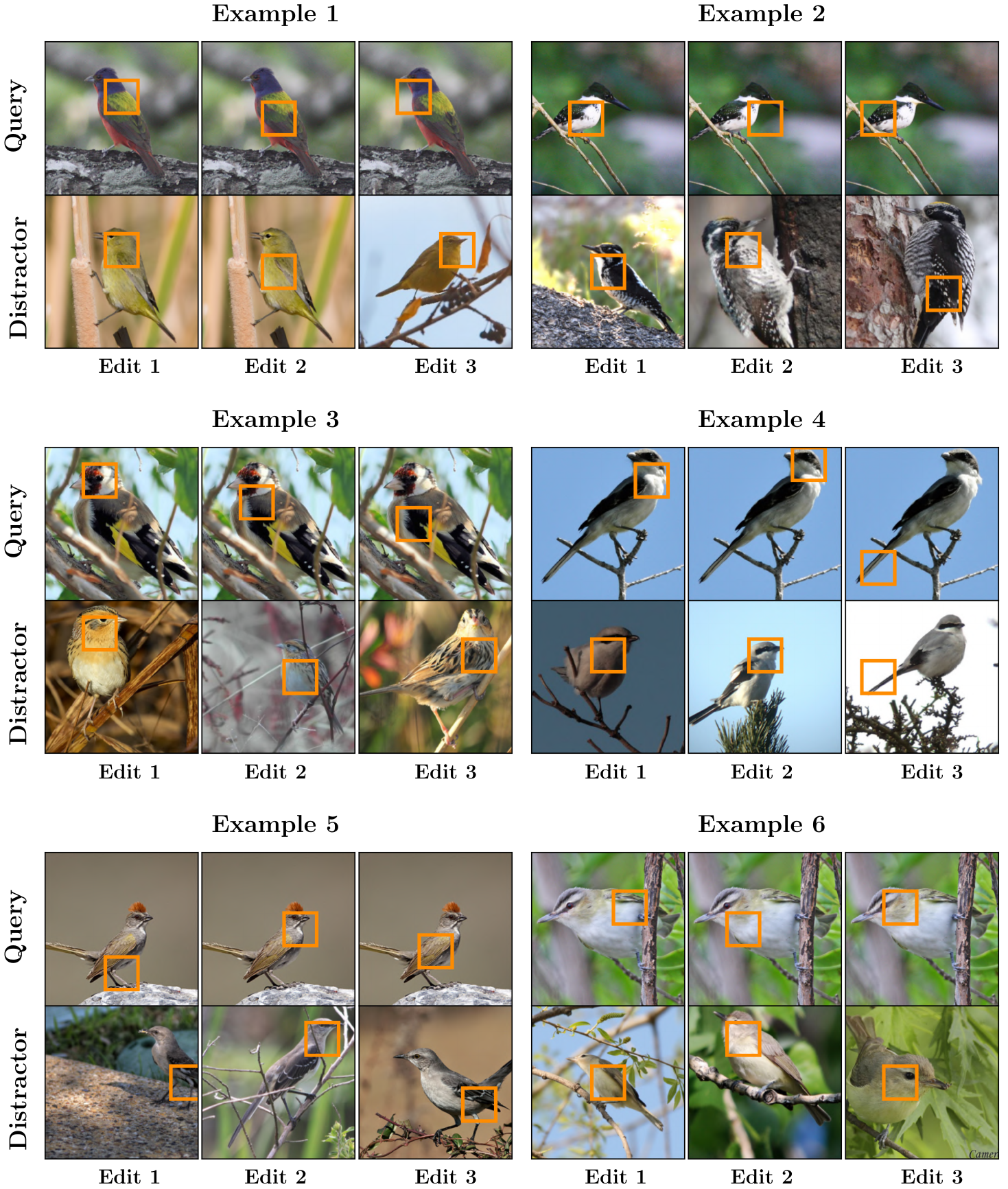}
    \caption{\textbf{Visualization of consecutive edits on CUB.} Our counterfactual explanations iteratively replace single cells until the model's decision changes. The figure highlights these consecutive edits in the query image and distractor image(s). To generate the figure, we select counterfactual explanation that use three edits.}
    \label{fig: suppl_multiple_edits}
\end{figure}

Recall that our method iteratively replaces cells between the query image and distractor image(s) until the model's decision changes. So far, we have only showed visualizations of the first cell edit. Figure~\ref{fig: suppl_multiple_edits} shows counterfactual explanations where we visualize all edits until the model's decision changes. For the purpose of visualization, we randomly selected counterfactual explanations that require three cell replacements. In each case, we observe that all edits select class-specific and semantically consistent image regions. In conclusion, our method achieves the desired result, not only for the first edit, but across all edits. 

\subsubsection{Visualization of failure cases.} 
\begin{figure}
    \centering
    \includegraphics[width=\textwidth]{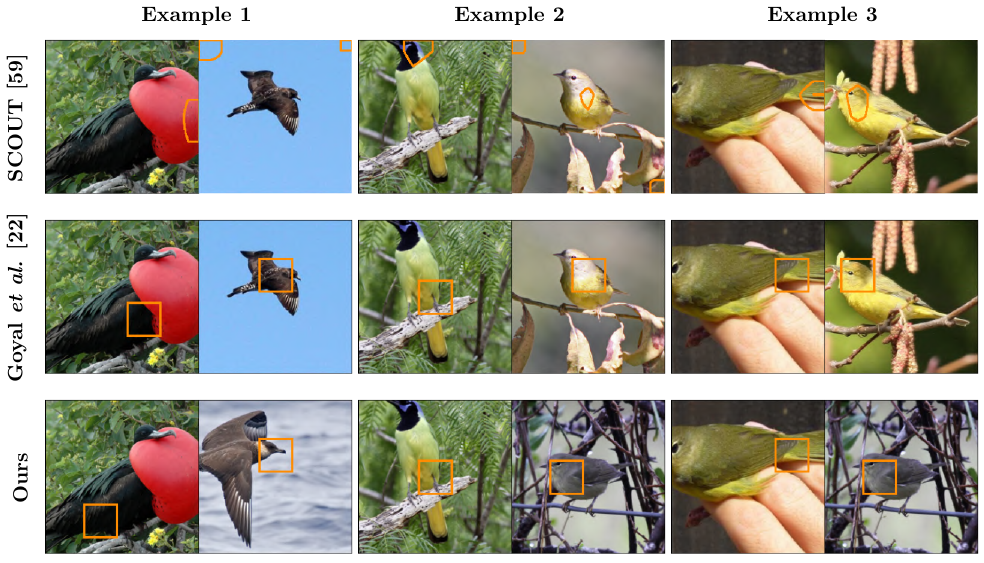}
    \caption{\textbf{Failure cases}. We show some failure cases, where our counterfactual explanations replace regions referring to different parts.}
    \label{fig: suppl_fails}
\end{figure}

Figure~\ref{fig: suppl_fails} shows three failure cases, where our counterfactual explanations replace regions referring to different bird parts. The examples were generated for a ResNet-50 classifier trained on the CUB dataset. We make the following observations. First, the failure cases seem to occur for (i) odd looking birds; e.g., the query bird in Example 1 looks quite different from other birds in the CUB dataset, and (ii) for query images where certain bird parts fall outside the image; e.g., the discriminative parts of the query bird in Example 2 and Example 3 fall outside the image. Second, we observe that other methods make similar mistakes. In conclusion, the failure cases seem caused by the challenging nature of the examples, rather than being a pitfall in our method.

\begin{figure}
    \centering
    \includegraphics[width=\textwidth]{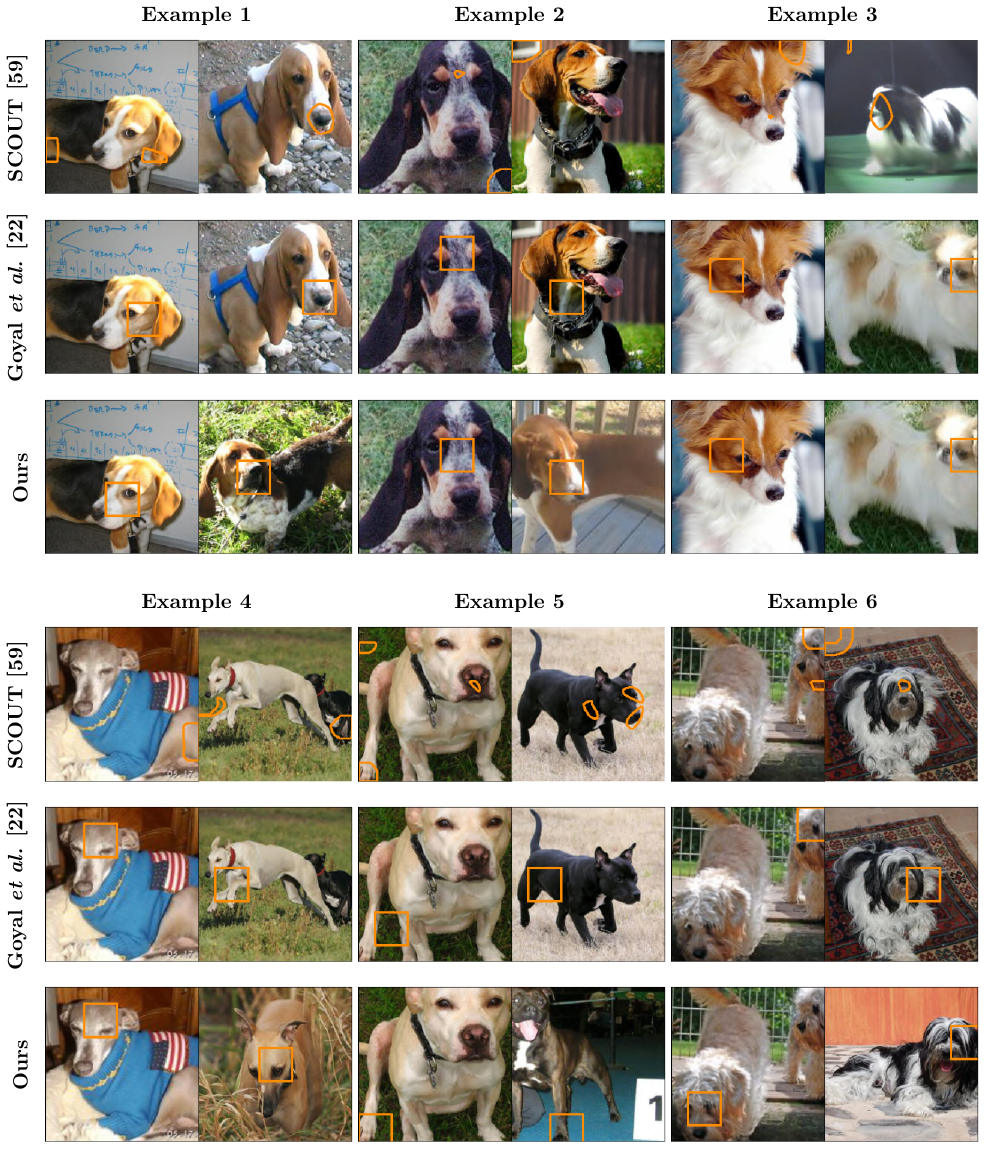}
        \caption{\textbf{Additional qualitative results on StanfordDogs~\cite{DOGS}.} We highlight the best edit in the query image (left) and distractor image (right).}
    \label{fig: suppl_dogs_res50}
\end{figure}

\clearpage

\section{Computational cost analysis}
\label{suppl: complexity}
In this section, we perform a complexity analysis of our approach. Additionally, we include a compute time analysis under the multi-distractor setup.

\subsubsection{Computational complexity.} We perform a back-of-the-envelope calculation of the number of multiply-add computations (MACs) in our framework. To simplify the analysis, we consider the \underline{computational cost of performing a single edit}. Recall from Sec.~\ref{sec: method} that the computational complexity of the classification loss ($\mathcal{L}_c$ in Eq.~\ref{eq: class_loss}) and semantic loss ($\mathcal{L}_s$ in Eq.~\ref{eq: similarity_loss}) can be summarized as:
\begin{equation}
    C_{\mathcal{L}_c} = 2 \cdot C_f + h^2 w^2 \cdot C_g
\end{equation}
\begin{equation}
    C_{\mathcal{L}_s} = 2 \cdot C_u + h^2 w^2 \cdot C_{\text{dot}}.
\end{equation}

We compute counterfactual explanations using the $7 \times 7 \times 512$ spatial features of the \texttt{max\_pooling2d\_5} layer in VGG-16~\cite{simonyan2014very}. The evaluation of the classification loss $\mathcal{L}_c$ uses $\num{318.8e9}$ MACs ($\approx 2 \cdot \num{15.4e9}$ for $f$ + $h^2w^2 \cdot \num{1.2e8}$ for $g$), while the semantic loss $\mathcal{L}_s$ computation uses $\num{8.2e9}$ MACs ($\approx 2 \cdot \num{4.1e9}$ for $u$ + $h^2w^2 \cdot \num{2.0e3}$ for the dot-product in the softmax $s$). We conclude that the classification loss is expensive to compute due to it's quadratic dependence on the number of cells $hw$ and the relatively high cost of evaluating the decision network $g(\cdot)$. In contrast, the semantic loss does not suffer from it's quadratic term because the dot-product operation is inexpensive to compute. In conclusion, $\mathcal{L}_s$ can be computed more efficiently compared to $\mathcal{L}_c$.

The pre-filtering operation from Sec.~\ref{subsec: method_multi_distractor} relies on the fast computation of the semantic similarity loss to realize a speed-up. For example, we select the top-10\% most similar cells ($k=0.1$) according to the semantic loss, and then only consider the classification loss for this subset of cells. This reduces the complexity of $\mathcal{L}_c$ to $\num{59.6e9}$ MACs ($\approx 2 \cdot \num{15.4e9}$ for $f$ + $\underline{k}h^2w^2 \cdot \num{1.2e8}$ for $g$). As a result, the overall computational cost is reduced from $\num{327e9}$ to $\num{67e9}$ MACs, meaning our framework holds a significant speed advantage over methods that compute $\mathcal{L}_c$ exhaustively~\cite{goyal2019counterfactual}.

This analysis does not consider the memory aspect of computing $\mathcal{L}_c$ and $\mathcal{L}_s$. It's worth noting that the similarity loss also holds an advantage in terms of memory usage compared to the classification loss. Specifically, in order to compute the classification loss for all $h^2w^2$ permutations, we need to construct all permutations in memory. This involves the allocation of $h^2w^2$ spatial cell matrices by replacing cells in $f(I)$ with cells from $f(I')$. In contrast, computing the similarity loss does not require to allocate $O(h^2w^2)$ extra memory as it does not involve replacing cells. Instead, the semantic loss operates directly on the spatial feature matrices of the auxiliary model, i.e., $u(I)$ and $u(I')$.

\begin{figure}[t]
\centering
\includegraphics[width=0.5\textwidth]{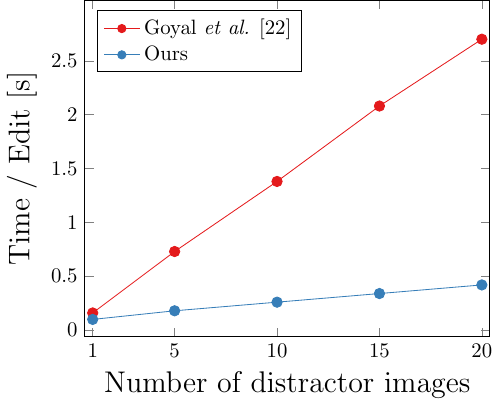}
\vspace{-1em}
\caption{Time analysis of multi-distractor setup.}
\label{fig: num_distractor_time}
\vspace{-2em}
\end{figure}

\noindent \textbf{Compute time analysis in a multi-distractor setup:} Figure~\ref{fig: num_distractor_time} reports the average computation time per edit (on a single V-100 GPU) as a function of the number of distractor images. We note that our method which includes a pre-filtering operation (Sec.~\ref{subsec: method_multi_distractor}) scales linearly with the number of distractor images and is about an order of magnitude faster compared to~\cite{goyal2019counterfactual}. 

\section{Attributes}
\label{suppl: attributes}

\subsection{Implementation details}
We provide additional implementation details of how we add natural language attribute information to the visual counterfactual explanations in Sec.~\ref{sec: attributes}. Recall, the classifier is a ResNet-50 model trained to identify bird species on CUB. The spatial feature extractor $f$ computes the $h \times w \times d$ spatial feature output of the last convolutional layer, and $g$ performs a global average pooling operation followed by a linear classifier. 

\subsubsection{Parts detector.} We reuse the CUB parts detector from Sec.~\ref{suppl: training}. The parts predictor is used to select the top-3 parts for the spatial cells that are being replaced in the counterfactual.

\subsubsection{Attribute classifiers.} We train linear classifiers to predict part-attributes on top of the average-pooled features from $f(\cdot)$. The part-attributes are derived from the attribute annotations used by~\cite{koh2020concept}. Specifically, we only use attributes that refer to parts for which we have keypoint locations. This results in 77 attributes in total. We train linear classifiers to predict the part-attributes via a multi-class cross-entropy loss. The training uses SGD with momentum $0.9$ and initial learning rate $0.04$. We use batches of size 64 and train for 100 epochs. The learning rate is decayed by 10 after 70 and 90 epochs. We use weight decay 1e-6. 

\subsubsection{Interpretable basis decomposition.} We perform the interpretable basis decomposition as follows. Consider a counterfactual that replaces a cell $i$ in $f(I)$ with a cell $i'$ from $f(I')$. First, we determine the attributes that should be used for the decomposition. To this end, we take the union of detected parts in cell $i$ and $i'$ first, and then select the attributes that are associated with the detected parts, e.g., if one of the parts is `wing' we select attributes like `has\_wing\_color::blue'. We then apply the algorithm from~\cite{zhou2018interpretable} to decompose the weights of the linear layer in $g$ in terms of the selected attribute classifiers. The decomposition is performed for the query $f(I)$ and counterfactual $f(I^*)$.

\subsubsection{Additional examples.} Figure~\ref{fig: attributes} shows additional examples, where our method succeeds in adding attribute information to our visual counterfactual explanations. In each case, the returned attribute belongs to class $c$ but not to class $c'$, or vice-versa. Thus, the returned attributes are discriminative of the classes $c$ and $c'$.

\begin{figure}
    \centering
    \includegraphics[width=\textwidth]{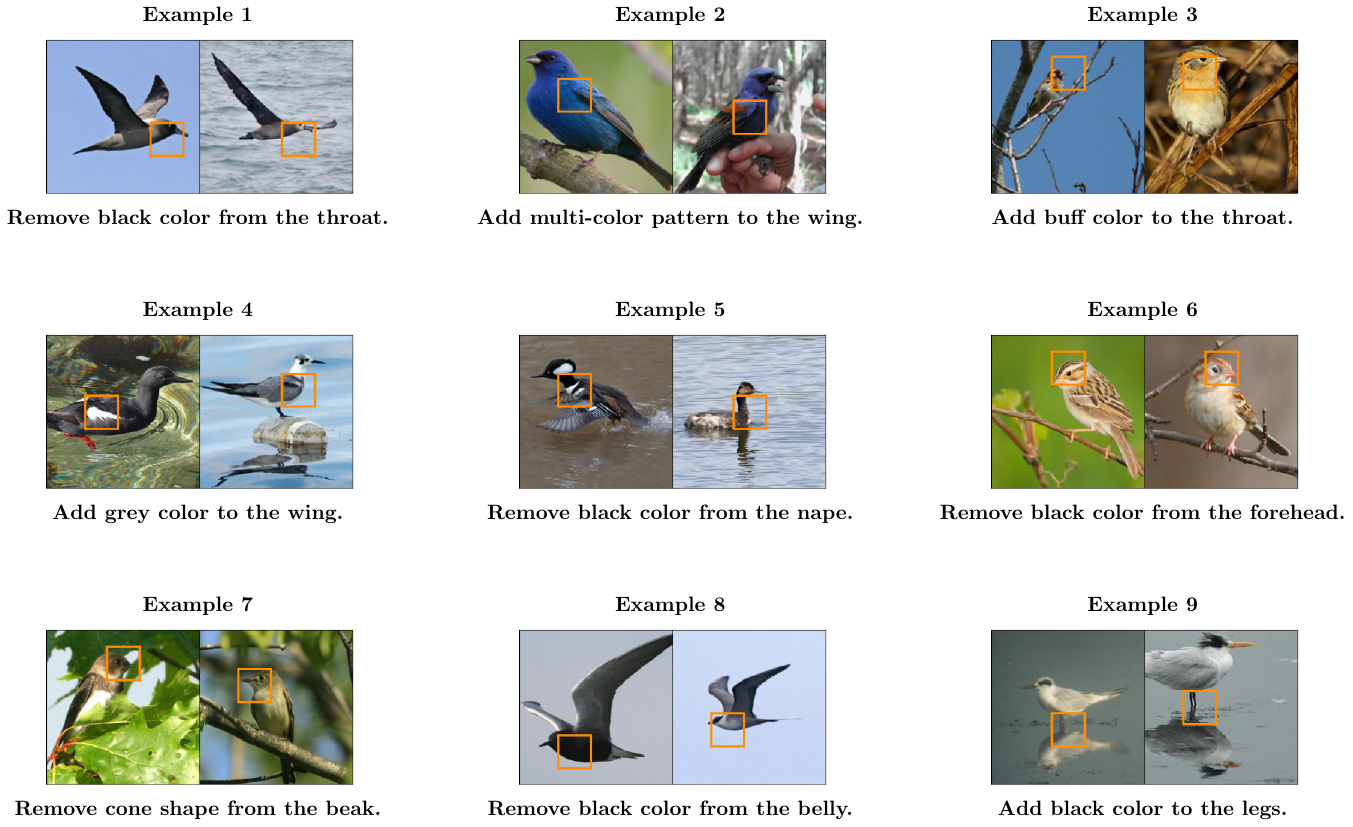}
    \caption{\textbf{Language-based counterfactuals.} We identify the attribute that is most important for chaenging the model's decision.}
    \label{fig: suppl_attributes}
\end{figure}

\clearpage
\section{Licenses}
\label{suppl: licenses}
We include the licenses for images from iNaturalist used in our visualizations.

\begin{table*}
\caption{\small{Authors and Creative Commons Copyright notice for images in Figure~\ref{fig: suppl_inat_birds_res50}.}}
\centering
\begin{minipage}{0.8\textwidth}
lauriekoepke: \href{http://creativecommons.org/licenses/by-nc/4.0/}{CC BY-NC 4.0}, 
walterflocke: \href{http://creativecommons.org/licenses/by-nc/4.0/}{CC BY-NC 4.0}, 
leo\_v: \href{http://creativecommons.org/licenses/by-nc/4.0/}{CC BY-NC 4.0}, 
Herbert Herbinia: \href{http://creativecommons.org/licenses/by-nc/4.0/}{CC BY-NC 4.0}, 
leptim: \href{http://creativecommons.org/licenses/by-nc/4.0/}{CC BY-NC 4.0}, 
leptim: \href{http://creativecommons.org/licenses/by-nc/4.0/}{CC BY-NC 4.0}, 
nanorca13: \href{http://creativecommons.org/licenses/by-nc/4.0/}{CC BY-NC 4.0}, 
toucan55: \href{http://creativecommons.org/licenses/by-nc/4.0/}{CC BY-NC 4.0}, 
Jim Brighton: \href{http://creativecommons.org/licenses/by-nc/4.0/}{CC BY-NC 4.0}, 
wildmouse3: \href{http://creativecommons.org/licenses/by-nc/4.0/}{CC BY-NC 4.0}, 
Will Richardson: \href{http://creativecommons.org/licenses/by-nc/4.0/}{CC BY-NC 4.0}, 
Amado: \href{http://creativecommons.org/licenses/by-nc/4.0/}{CC BY-NC 4.0}, 
Esteban Munguia: \href{http://creativecommons.org/licenses/by-nc/4.0/}{CC BY-NC 4.0}, 
jamesbeat: \href{http://creativecommons.org/licenses/by-nc/4.0/}{CC BY-NC 4.0}, 
jamesbeat: \href{http://creativecommons.org/licenses/by-nc/4.0/}{CC BY-NC 4.0}, 
N. Mahathi: \href{http://creativecommons.org/licenses/by-nc/4.0/}{CC BY-NC 4.0}, 
ddun: \href{http://creativecommons.org/licenses/by-nc/4.0/}{CC BY-NC 4.0}, 
John G. Phillips: \href{http://creativecommons.org/licenses/by-nc/4.0/}{CC BY-NC 4.0}.
\end{minipage}
\end{table*}

\begin{table*}
\caption{\small{Author and Creative Commons Copyright notice for images in Figure~\ref{fig: suppl_inat_other_res50}.}}
\centering
\begin{minipage}{0.8\textwidth}
Daniel George: \href{http://creativecommons.org/licenses/by-nc/4.0/}{CC BY-NC 4.0}, 
thehaplesshiker: \href{http://creativecommons.org/licenses/by-nc/4.0/}{CC BY-NC 4.0}, 
tiyumq: \href{http://creativecommons.org/licenses/by-nc/4.0/}{CC BY-NC 4.0}, 
Rohit Chakravarty: \href{http://creativecommons.org/licenses/by-nc/4.0/}{CC BY-NC 4.0}, 
ungerlord: \href{http://creativecommons.org/licenses/by-nc/4.0/}{CC BY-NC 4.0}, 
dushenkov: \href{http://creativecommons.org/licenses/by-nc/4.0/}{CC BY-NC 4.0}, 
jenhenlo: \href{http://creativecommons.org/licenses/by-nc/4.0/}{CC BY-NC 4.0}, 
amniotasmarinos6: \href{http://creativecommons.org/licenses/by-nc/4.0/}{CC BY-NC 4.0}, 
Lawrence Troup: \href{http://creativecommons.org/licenses/by-nc/4.0/}{CC BY-NC 4.0}, 
Andrew Deacon: \href{http://creativecommons.org/licenses/by-nc/4.0/}{CC BY-NC 4.0}, 
Torbjorn von Strokirch: \href{http://creativecommons.org/licenses/by-nc/4.0/}{CC BY-NC 4.0},
\begin{CJK*}{UTF8}{gbsn}
金翼白眉: \href{http://creativecommons.org/licenses/by-nc/4.0/}{CC BY-NC 4.0}, 
\end{CJK*}
Andrés Matos: \href{http://creativecommons.org/licenses/by-nc/4.0/}{CC BY-NC 4.0}, 
Christoph Moning: \href{http://creativecommons.org/licenses/by-nc/4.0/}{CC BY-NC 4.0},
asandlermd: \href{http://creativecommons.org/licenses/by-nc/4.0/}{CC BY-NC 4.0}, 
Eric Giles: \href{http://creativecommons.org/licenses/by-nc/4.0/}{CC BY-NC 4.0}, 
Paul Cools: \href{http://creativecommons.org/licenses/by-nc/4.0/}{CC BY-NC 4.0}, 
Laura Kimberly: \href{http://creativecommons.org/licenses/by-nc/4.0/}{CC BY-NC 4.0}, 
sterling: \href{http://creativecommons.org/licenses/by/4.0/}{CC BY 4.0}, 
elisabraz: \href{http://creativecommons.org/licenses/by-nc/4.0/}{CC BY-NC 4.0}, 
Tom Warnert: \href{http://creativecommons.org/licenses/by-nc/4.0/}{CC BY-NC 4.0}, 
Don Loarie: \href{http://creativecommons.org/licenses/by/4.0/}{CC BY 4.0}, 
Marlo Perdicas: \href{http://creativecommons.org/licenses/by/4.0/}{CC BY 4.0}, 
luca tringali: \href{http://creativecommons.org/licenses/by-nc/4.0/}{CC BY-NC 4.0},
pmmullins: \href{http://creativecommons.org/licenses/by-nc/4.0/}{CC BY-NC 4.0}, 
Erik Schlogl: \href{http://creativecommons.org/licenses/by-nc/4.0/}{CC BY-NC 4.0}, 
sea-kangaroo: \href{http://creativecommons.org/licenses/by-nc-nd/4.0/}{CC BY-NC-ND 4.0}, 
Jason Grant: \href{http://creativecommons.org/licenses/by-nc/4.0/}{CC BY-NC 4.0}, 
Donald Hobern: \href{http://creativecommons.org/licenses/by/4.0/}{CC BY 4.0}, 
Richard Ling: \href{http://creativecommons.org/licenses/by-nc-nd/4.0/}{CC BY-NC-ND 4.0}, 
BeachBumAgg: \href{http://creativecommons.org/licenses/by-nc/4.0/}{CC BY-NC 4.0}, 
Tony Strazzari: \href{http://creativecommons.org/licenses/by-nc/4.0/}{CC BY-NC 4.0}.
\end{minipage}
\end{table*}

\end{document}